\documentclass{article}

\usepackage{times}
\usepackage{graphicx} 
\usepackage{subfigure} 

\usepackage{natbib}

\usepackage{algorithm}
\usepackage{algorithmic}

\usepackage{hyperref}

\usepackage[accepted]{icml2016}

\usepackage{url}

\usepackage{amsmath,amssymb}

\usepackage{defs}

\icmltitlerunning{Scalable Discrete Sampling as a Multi-Armed Bandit Problem}

\begin{document} 

\twocolumn[
\icmltitle{Scalable Discrete Sampling as a Multi-Armed Bandit Problem}

\icmlauthor{Yutian Chen}{yutian.chen@eng.cam.ac.uk}
\icmladdress{Department of Engineering, University of Cambridge, Cambridge CB2 1PZ, UK}
\icmlauthor{Zoubin Ghahramani}{zoubin@eng.cam.ac.uk}
\icmladdress{Department of Engineering, University of Cambridge, Cambridge CB2 1PZ, UK\\
Alan Turing Institute, 96 Euston Road, London NW1 2DB, UK}

\icmlkeywords{discrete sampling, large scale, subsampling, MCMC, Multi-Armed Bandit}

\vskip 0.3in
]

\begin{abstract} 
Drawing a sample from a discrete distribution is one of the building components for Monte Carlo methods. Like other sampling algorithms, discrete sampling suffers from the high computational burden in large-scale inference problems. We study the problem of sampling a discrete random variable with a high degree of dependency that is typical in large-scale Bayesian inference and graphical models, and propose an efficient approximate solution with a subsampling approach. We make a novel connection between the discrete sampling and Multi-Armed Bandits problems with a finite reward population and provide three algorithms with theoretical guarantees. Empirical evaluations show the robustness and efficiency of the approximate algorithms in both synthetic and real-world large-scale problems.
\end{abstract} 

\section{Introduction}

Sampling a random variable from a discrete (conditional) distribution is one of the core operations in Monte Carlo methods. It is an ubiquitous and often necessary component for inference algorithms such as Gibbs sampling and particle filtering. Applying discrete sampling for large-scale problems has been a challenging task like other Monte Carlo algorithms due to the high computational burden. Various approaches have been proposed to address different dimensions of ``large scales". For example, distributed algorithms have been used to sample a model with a large number of discrete variables \cite{newman2009distributed,bratieres2010parallelihmm,bicksonkddcup2011workshop}, smart transition kernels were described for Markov chain Monte Carlo (MCMC) algorithms to sample efficiently a single variable with a large or even infinite state space \cite{fastlda2014,kalli2011slice}. This paper is focused on another dimension of the ``large-scales" where the variable to sample has a large degree of statistical dependency.

Consider a random variable with a finite domain $X \in \cX$ and a distribution in the following form
\begin{align}
p(X=x) \propto \tilde{p}(X=x), \text{ with } \tilde{p}(X=x) = f_0(x) \prod_{n=1}^N f_n(x),\label{eq:tilde_p}
\end{align}
where $f_n$ can be any function of $x$. Such distribution occurs frequently in machine learning problems. For example, in Bayesian inference for a model with parameter $X$ and $N$ i.i.d.\ observations $\mathcal{D}=\{\by_n\}_{n=1}^N$, the posterior distribution of $X$ depends on all the observations when sufficient statistics is not available. The unnormalized posterior distribution can be written as $\tilde{p}(X|\mathcal{D})=p(X)\prod_{i=1}^N p(\by_i|X)$. In undirected graphical model inference problems where a node $X_i$ appears in $N$ potential functions, the distribution of $X_i$ depends on the value of all of the $N$ functions. The unnormalized conditional distribution is $\tilde{p}(X_i|\bx_{-i})=\prod_{n=1}^N \phi_n(X_i, \bx_{-i})$, where $\bx_{-i}$ denotes the value of all the other nodes in the graph and $\phi_n$ denotes a potential function that includes $X_i$ in the scope. In this paper we study how to sample a discrete random variable $X$ in a manner that is scalable in $N$.

A common approach to address the big data problem is divide-and-conquer that uses parallel or distributed computing resources to process data in parallel and then synchronize the results periodically or merely once in the end  \cite{scott2013bayes,medlar2013swiftlink,XuTehZhu2014a}.

An orthogonal approach has been studied for the Metropolis-Hastings (MH) algorithm in a general state space by running a sampler with subsampled data. This approach can be combined easily with the distributed computing idea for even better scalability \citep[e.g.][]{ahn2015large}.

\citet{maclaurin2015firefly} introduced an MH algorithm in an augmented state space that could achieve higher efficiency than the standard MH by processing only a subset of active data every iteration while still preserving the correct stationary distribution. But the introduction of auxiliary variables might also slow down the overall mixing rate in the augmented space.

Approximate MH algorithms have been proposed in the subsampling approach with high scalability. The stochastic gradient Langevin dynamics (SGLD) \cite{welling2011bayesian} and its extensions \cite{ahn2012bayesian,chen2014stochastic,ding2014bayesian} introduced efficient proposal distributions based on subsampled data. Approximate algorithms induce bias in the stationary distribution of the Markov chain. But given a fixed amount of runtime they could reduce the expected error in the Monte Carlo estimate via a proper trade-off between variance and bias by mixing faster w.r.t.\ the runtime. This is particularly important for large-scale learning problems when the runtime is one of the limiting factors for generalization performance \cite{BottouBousquet08}. However, the stochastic gradient MCMC approach usually skips the rejection step in order to obtain sublinear time complexity and the induced bias is very hard to estimate or control.

Another line of research on approximate subsampled MH algorithms does not ignore the rejection step but controls the error with an approximate rejection step based on a subset of data \cite{korattikara2013austerity,bardenet2014towards}. The bias can thus be better controlled \cite{mitrophanov2005sensitivity,pillai2014ergodicity}. That idea has also been extended to slice sampling \cite{dubois2014approximate} and Gibbs for binary variables \cite{korattikara2013austerity}.

In this paper we follow the last line of research and propose a novel approximate sampling algorithm to improve the scalability of sampling \textit{discrete} distributions. We first reformulate the problem in Eq.~\ref{eq:tilde_p} as a Multi-Armed Bandit (MAB) problem with a finite reward population via the Gumbel-Max trick \cite{PaYu11a}, and then propose three algorithms with theoretical guarantees on the approximation error and an upper bound of $N|\cX|$ on the sample size. This is to our knowledge the first attempt to address discrete sampling problem with a large number of dependencies and our work will likely contribute to a more complete library of scalable MCMC algorithms. Moreover, the racing algorithm in Sec.~\ref{sec:racing} provides a unified framework for subsampling-based discrete sampling, MH \cite{korattikara2013austerity,bardenet2014towards} and slice sampling \cite{dubois2014approximate} algorithms as discussed in Sec.~\ref{sec:related_work}. We also show in the experiments that our algorithm can be combined straightforwardly with stochastic gradient MCMC to achieve both high efficiency and controlled bias. Lastly, the proposed algorithms also deserve their own interest for MAB problems under this particular setting.

We first review an alternative way of drawing discrete variables and build a connection with MABs in Sec.~\ref{sec:disc_sampling}, then propose three algorithms in Sec.~\ref{sec:algs}. We discuss related work in Sec.~\ref{sec:related_work} and evaluate the proposed algorithms on both synthetic data and real-world problems of Bayesian inference and graphical model inference in Sec.~\ref{sec:experiments}. Particularly, we show how our proposed sampler can be combined conveniently as a building component with other subsampling sampler for a hierarchical Bayesian model. Sec.~\ref{sec:discussion} concludes the paper with a discussion.

\section{Approximate Discrete Sampling}\label{sec:disc_sampling}
\subsection{Discrete Sampling as an Optimization Problem}
The common procedure to sample $X$ from a discrete domain $\mathcal{X} = \{1,2,\dots,D\}$ is to first normalize $\tilde{p}(X)$ and compute the CDF $F(X=x)=\sum_{i=1}^x p(X=i)$. Then draw a uniform random variable $u\sim \mathrm{Uniform}(0,1]$, and find $x$ that satisfies
$F(x-1) < u \leq F(x)$.
This procedure requires computing the sum of all the unnormalized probabilities. For $\tilde{p}$ in the form of Eq.~\ref{eq:tilde_p} this is $\mathcal{O}(ND)$.

An alternative procedure is to first draw $D$ i.i.d.\ samples from the standard Gumbel distribution%
\footnote{The Gumbel distribution is used to model the maximum extreme value distribution. If a random variable $Z \sim \mathrm{Exp(1)}$, then $-\log(Z) \sim \mathrm{Gumbel}(0,1)$. $\eps$ can be easily drawn as $-\log(-\log(u))$ with $u\sim \mathrm{U}[0,1]$.}
$\eps_i \sim \mathrm{Gumbel}(0,1)$, and then solve the following optimization problem:
\begin{equation}
x = \argmax_{i\in \mathcal{X}} \log \tilde{p}(i) + \eps_i. \label{eq:disc_gumbel}
\end{equation}
It is shown in \citet{kuzmin2005optimum} that $x$ follows the distribution $p(X)$. With this method after drawing random variables that do not depend on $\tilde{p}$, we turn a random sampling problem to an optimization problem. While the computational complexity is the same to draw an \textit{exact} sample, 
an \textit{approximate} algorithm may potentially save computations by avoiding computing accurate values of $\tilde{p}(X=x)$ when $x$ is considered unlikely to be the maximum as discussed next.

\subsection{Approximate Discrete Sampling as a Multi-Armed Bandits Problem}
In a Multi-Armed Bandit (MAB) problem, the $i$'th bandit is a slot machine with an arm, which when pulled generates an i.i.d.\ reward $l_i$ from a distribution associated with that arm with an unknown mean $\mu_i$. The optimal arm identification problem for MABs \cite{bechhofer1958sequential,paulson1964sequential} in the fixed confidence setting is to find the arm with the highest mean reward with a confidence $1-\de$ using as few pulls as possible.

Under the assumption of Eq.~\ref{eq:tilde_p}, the solution in Eq.~\ref{eq:disc_gumbel} can be expressed as
\begin{align}
x &= \argmax_{i\in \mathcal{X}} \sum_{n=1}^N \log f_n(i) + \log f_0(i) + \eps_i \nn\\
  &= \argmax_{i\in \mathcal{X}} \sum_{n=1}^N \underbrace{\left(\log f_n(i) + \frac{1}{N}\left(\log f_0(i) + \eps_i\right)\right)}_{\defeq l_{i,n}} \nn
\end{align}
\begin{align}
  &= \argmax_{i\in \mathcal{X}} \frac{1}{N} \sum_{n=1}^N l_{i,n} = \argmax_{i\in \mathcal{X}} \mathbb{E}_{l_i \sim \text{Uniform}(\mathcal{L}_i)} [l_i] \nn\\
  &\defeq \argmax_{i\in \mathcal{X}} \mu_i \label{eq:x_mab}
\end{align}
where $\mathcal{L}_i \defeq \{l_{i,1},l_{i,2},\dots,l_{i,N}\}$. After drawing $D$ Gumbel variables $\eps_i$, we turn the discrete sampling problem into the optimal arm identification problem in MABs where the reward $l_i$ is uniformly sampled from a finite population $\mathcal{L}_i$. 
An approximate algorithm that solves the problem with a fixed confidence may avoid drawing all the rewards from an obviously sub-optimal arm and save computations. 
We show the induced bias in the sample distribution as follows with the proof in Appx.~\ref{sec:proof_bound}.
\begin{proposition}\label{prop:bound}
If an algorithm solves (\ref{eq:disc_gumbel}) exactly
 with a probability at least $1-\de$ for any value of $\beps$, the total variation between the sample distribution $\hat{p}$ and the true distribution is bounded by
\begin{equation}
\|\hat{p}(X) - p(X)\|_{\mathrm{TV}} \leq \de \label{eq:sample_distr_error}
\end{equation}
\end{proposition}
When applied in the MCMC framework as a transition kernel, we can apply immediately the theories in \citet{mitrophanov2005sensitivity,pillai2014ergodicity} to show that the approximate Markov chain satisfies uniform ergodicity under regular conditions and the analysis of convergence rate are readily available under various assumptions. So the discrete sampling problem of this paper reduces to finding a good MAB algorithm for Eq.~\ref{eq:disc_gumbel} in our problem setting.

\section{Algorithms for MABs with a Finite Population and Fixed Confidence}\label{sec:algs}

The key difference of our problem from the regular MABs is that our rewards are generated from a finite population while regular MABs assume i.i.d.\ rewards. Because one can obtain the exact mean by sampling all the $N$ values $l_{i,n}$ for arm $i$ \textit{without replacement}, a good algorithm should pull no more than $N$ times for each arm regardless of the mean gap between arms. We introduce three algorithms in this section whose sample complexity is upper bounded by $O(ND)$ in the worst case and can be very efficient when the mean gap is large.

\subsection{Notations}
The iteration of an algorithm is indexed by $t$. We denote the entire index set with $[N]=\{1,2,\dots,N\}$, the sampled set of reward indices up to $t$'th iteration from arm $i$ with $\cN_i^{(t)}\subseteq [N]$, and the corresponding number of sampled rewards with $T_i^{(t)}$. We define the estimated mean for $i$'th arm with $\hat{\mu}_i^{(t)}\defeq \frac{1}{|\cN_i^{(t)}|} \sum_{n\in \cN_i^{(t)}} l_{i,n}$, the natural variance (biased) estimate with $(\hat{\sg}_i^{(t)})^2 \defeq \frac{1}{|\cN_i^{(t)}|} \sum_{n\in \cN_i^{(t)}} (l_{i,n}-\hat{\mu}_i^{(t)})^2$, the variance estimate of the mean gap between two arm with $(\hat{\sg}_{i,j}^{(t)})^2 \defeq \frac{1}{|\cN_i^{(t)}|} \sum_{n\in \cN_i^{(t)}} ((l_{i,n} - l_{j,n}) - (\hat{\mu}_i^{(t)}-\hat{\mu}_j^{(t)}))^2$ (defined only when $\cN_i^{(t)}=\cN_j^{(t)}$), the bound of the reward value $C_i\defeq \max_{n,n'} \{l_{i,n} - l_{i,n'}\}$. The subscripts and superscripts may be dropped for notational simplicity when the meaning is clear from the context.

\subsection{Adapted lil'UCB}
We first study one of the state-of-the-art algorithms for fixed-confidence optimal arm identification problem and adjust it for the finite population setting. The lil'UCB algorithm \cite{jamieson2014lil} maintains an upper confidence bound (UCB) of $\mu_i$ that is inspired by the law of the iterated logarithm (LIL) for every arm. At each iteration, it draws a single sample from the arm with the highest bound and updates it. The algorithm terminates when some arm is sampled much more often than all the other arms. We refer readers to Fig.~1 of \citet{jamieson2014lil} for details. The time complexity for $t$ iterations is $\mathcal{O}(\log(D)t)$.
It was shown in \citet{jamieson2014lil} that lil'UCB achieved the optimal sample complexity up to constants.

However, lil'UCB requires i.i.d.\ rewards for each arm $i$, that is, sampled with replacement from $\cL_i$. Therefore, the total number of samples $t$ is unbounded and could be $\gg ND$ when the means are close to each other. We adapt lil'UCB for our problem with the following modifications:
\begin{enumerate}
\item Samples $l_{i,n}$ \textit{without replacement} for each arm but keep different arms independent.
\item When $T_i^{(t)} = N$ for some arm $i$, the estimate $\hat{\mu}_i^{(t)}$ becomes exact. So set its UCB to $\hat{\mu}_i^{(t)}$.
\item The algorithm terminates either with the original stopping criterion or when the arm with the highest upper bound has an exact mean estimate, whichever comes first.
\end{enumerate}
The adapted algorithm satisfies all the theoretical guarantees in Thm.~2 of \citet{jamieson2014lil} with additional properties as shown in the following proposition with proof in Appx.~\ref{sec:proof_lilucb}.
\begin{proposition}\label{prop:lil'ucb}
Theorem 2 of \citet{jamieson2014lil} holds for the adapted lil'UCB algorithm. Moreover $T_i^{(t)} \leq N, \forall i, t$. Therefore, when the algorithm terminates, $t = \sum_{i\in \cX} T_i^{(t)} \leq ND$.
\end{proposition}
Notice that Thm.~2 of \citet{jamieson2014lil} shows that $t$ scales roughly as $O(1/\Delta^2)$ with $\Delta$ being the mean gap and therefore $t \ll ND$ when the gap is large.

\subsection{Racing Algorithm for a Finite Population}\label{sec:racing}
When rewards are sampled without replacement, the negative correlation between rewards would generally improve the convergence of $\hat{\mu}_i$. Unfortunately, the bound in lil'UCB ignores the negative correlation when $T_i^{(t)}<N$ even with the adaptations. We introduce a new family of racing algorithms \cite{maron1993hoeffding} that takes advantage of the finite population setting as shown in Alg.~\ref{alg:racing}. The choice of the uncertainty bound function $G$ differentiates specific algorithms and two examples will be discussed in the following sections.

Alg.~\ref{alg:racing} maintains a set of candidate set $\cD$ initialized with all arms. At iteration $t$, a shared mini-batch of $m^{(t)}$ indices are drawn w/o replacement for all survived arms in $\cD$.
Then the uncertainty bound $G$ is used to eliminate sub-optimal arms with a given confidence. The algorithm stops when only one arm remains. We require for $m^{(t)}$ that
the total number of sampled indices $T^{(t^*)}=\sum_{t=1}^{t^*} m^{(t)}$ equals $N$ at the last iteration $t^*$.
Particularly, we take a doubling schedule $T^{(t)}=2 T^{(t-1)}$ (so $t^*=\lceil\log_2\frac{N}{m^{(1)}}\rceil+1$) and leave $m^{(1)}$ as a free parameter. We also require $G(\cdot,T,\cdot,\cdot) = 0$ whenever $T=N$ so that Alg.~\ref{alg:racing} always stops within $t^*$ iterations. The computational complexity for $t$ iterations is $\mathcal{O}(DT^{(t)})$ with the marginal estimate $\hat{\sg}_i$ and $\mathcal{O}(D^2 T^{(t)})$ with the pairwise estimate $\hat{\sg}_{i,j}$. The former version is more efficient than the latter when $D$ is large at the price of a looser bound.

\begin{proposition}\label{prop:racing}
If $G$ satisfies 
\begin{equation}
\cE \defeq P(\exists t < t^*, \hat{\mu}^{(t)} - \mu > G(\de, T^{(t)}, \hat{\sg}^{(t)}, C)) \leq \de,\label{eq:condition_G}
\end{equation}
for any $\de \in (0, 1)$ with a probability at least $1-\de$, Alg.~\ref{alg:racing} returns the optimal arm with at most $ND$ samples.
\end{proposition}
The proof is provided in Appx.~\ref{sec:proof_racing}.
Unlike adapted lil'UCB, Racing draws a shared set of sample indices among all the arms and could provide a tighter bound with pairwise variance estimates $\hat{\sg}_{i,j}$ when there is positive correlation, a typical case in Bayesian inference problems.

\begin{algorithm}[t]
\caption{Racing Algorithm with a Finite Reward Population}
\label{alg:racing}
\begin{algorithmic}
\INPUT Number of arms $D$, population size $N$, mini-batch sizes $\{m^{(t)}\}_{t=1}^{t^*}$, confidence level $1-\de$, uncertainty bound function $G(\de, T, \hat{\sg}, C)$, range of samples $C_i$ (optional).
\STATE $t \gets 0$, $T \gets 0$, $\cD \gets \{1, 2, \dots, D\}$, $\cN \gets \emptyset$
\WHILE{$|\cD| > 1$}
  \STATE $t \gets t + 1$
  \STATE Sample w/o replacement $m^{(t)}$ indices $\cM \subseteq [N]\backslash \cN$, and set $\cN \gets \cN \cup \cM$, $T \gets T + m^{(t)}$
  \STATE Compute $l_{i,n}, \forall i\in \cD, n\in \cM$, and update $\hat{\mu}_i$ and $\hat{\sg}_i$ (or $\hat{\sg}_{i,j}$), $\forall i\in \cD$.
  \STATE Find the best arm $x \gets \argmax_{i\in \cD} \hat{\mu}_i$
  \STATE Eliminate sub-optimal arms when the estimated gap is large $\cD \gets \cD \backslash \{i: \hat{\mu}_x - \hat{\mu}_i > G(\frac{\de}{D}, T, \hat{\sg}_x, C_x) + G(\frac{\de}{D}, T, \hat{\sg}_i, C_i)\}$ (or $\cD \gets \cD \backslash \{i: \hat{\mu}_x - \hat{\mu}_i > G(\frac{\de}{D-1}, T, \hat{\sg}_{x,i}), C_x + C_i\}$)
\ENDWHILE
\OUTPUT $\cD$
\end{algorithmic}
\end{algorithm}

\subsubsection{Racing with Serfling Concentration bounds for $G$}\label{sec:racing-EBS}
\citet{serfling1974} studied the concentration inequalities of sampling without replacement and obtained an improved Hoeffding bound. \citet{bardenet2013concentration} extended the work and provided an empirical Bernstein-Serfling bound that was later used for the subsampling-based MH algorithm \cite{bardenet2014towards}: for any $\delta \in (0, 1]$ and any $n \leq N$, with probability $1 - \delta$, it holds that
\begin{align}
&\hat{\mu}_n - \mu \leq \hat{\sigma}_n \sqrt{\frac{2\rho_n \log(5/\de)}{n}} + \frac{\kappa C \log(5/\de)}{n} \nn\\
&\defeq B_{\text{EBS}}(\de, n, \hat{\sg}_n, C)
\end{align}
where $\kappa = \frac{7}{3} + \frac{3}{\sqrt{2}}$, and
$
\rho_n = \left\{\begin{array}{ll}
1 - \pi_{n-1} & \mbox{if }n \leq N/2 \\
(1 - \pi_n)(1 + \frac{1}{n}) & \mbox{if }n > N/2
\end{array} \right.\label{eq:empirical-Bernstein-Serfling}
$,
with $\pi_n \defeq \dfrac{n}{N}$.
The extra term $\rho_n$ that is missing in regular empirical Bernstein bounds reduces the bound significantly when $n$ is close to $N$. 
We set $m^{(1)}=2$ in Alg.~\ref{alg:racing} to provide a valid $\hat{\sg}^{(t)}$ for any $t$ and set the uncertain bound $G$ with the empirical Bernstein-Serfling (EBS) bounds as
\begin{equation}
G_{\text{EBS}}(\de, T, \hat{\sg}, C) = B_{\text{EBS}}\left(\frac{\de}{t^*-1}, T, \hat{\sg}, C\right)\label{eq:G_ebs}
\end{equation}
It is trivial to prove that $G_{\text{EBS}}$ satisfies the condition in Eq.~\ref{eq:condition_G} using a union bound over $t<t^*$.

\subsubsection{Racing with a Normal Assumption for $G$}\label{sec:racing-normal}
The concentration bounds often give a conservative strategy as they assume an arbitrary bounded reward distribution. When the number of drawn samples is large, the central limit theorem suggests that $\hat{\mu}^{(t)}$ follows approximately a Gaussian distribution. \citet{korattikara2013austerity} made such an assumption and obtained a tighter bound. We first provide an immediate corollary of Prop.~2 in Appx.~A of \citet{korattikara2013austerity}.
\begin{corollary}
Let $\hat{\mu}^{(t)}_{\mathrm{unit}}, t=1,2,\dots,t^*$ be the estimated means using sampling without replacement from any finite population with mean $\mu$ and unit variance. The joint normal random variables $\tilde{\mu}^{(t)}$ that match the mean and covariance matrix with $\hat{\mu}^{(t)}_{\mathrm{unit}}$ follow a Gaussian random walk process as
\begin{equation}
p_\mu(\tilde{\mu}^{(t)}|\tilde{\mu}^{(1)}, \dots, \tilde{\mu}^{(t-1)}) = \mathcal{N}(m_t(\tilde{\mu}^{(t-1)}), S_t) \label{eq:random_walk}
\end{equation}
where
$
m_t = \mu + A_t (\tilde{\mu}_{t-1}-\mu),
S_t = \frac{B_t}{T^{(t)}}\left(1 - \frac{T^{(t)}-1}{N-1}\right)
$
,
$
A_t = \frac{\pi_{t-1}(1-\pi_{t})}{\pi_{t}(1-\pi_{t-1})}
$,
$
B_t = \frac{\pi_{t} - \pi_{t-1}}{\pi_{t} (1-\pi_{t-1})}
$
with $\pi_t$ short for $\pi_{T^{(t)}}$.
\end{corollary}
\begin{remark}
The marginal distribution $p(\tilde{\mu}^{(t)})=\mathcal{N}\left(\mu, \frac{1}{T^{(t)}}\left(1 - \frac{T^{(t)}-1}{N-1}\right)\right)$ where the variance approaches 0 when $T^{(t)}\rightarrow N$.
\end{remark}
\begin{assumption}\label{assum:normal}
When $T^{(t)} \gg 1, \forall t$, we assume $\hat{\sg}^{(t)} \approx \sg$ and the central limit theorem suggests that the joint distribution of $\hat{\mu}^{(t)}/\hat{\sg}^{(t)}$ can be approximated by the joint distribution of $\tilde{\mu}^{(t)}$.
\end{assumption}
With the normal assumption, we choose the uncertainty bound $G$ in the following form
\begin{equation}
G_{\mathrm{Normal}}(\de, T, \hat{\sg}) = \frac{\hat{\sg}}{\sqrt{T}} \left(1-\frac{T-1}{N-1}\right)^{1/2} B_{\mathrm{Normal}} \label{eq:G_normal}
\end{equation}
Intuitively we use a constant confidence level, $\Phi(B_{\mathrm{Normal}})$, for all marginal distributions of $\hat{\mu}^{(t)}$ over $t$ where $\Phi(\cdot)$ is the CDF of the standard normal.
To choose the constant $B_{\mathrm{Normal}}$, we plug $G_{\mathrm{Normal}}$ into the condition for G in Eq.~\ref{eq:condition_G} and apply the normal distribution (\ref{eq:random_walk}) to solve the univariate equation $\cE(B)=\de$.
This way of computing $G$ gives a tighter bound than applying the union bound across $t$ as in the previous section because it takes into account the correlation of mean estimates across iterations.
Appx.~\ref{sec:table_B} provides a lookup table and a plot of $B_{\mathrm{Normal}}(\de)=\cE^{-1}(\de)$. Notice that $B_{\mathrm{Normal}}$ only needs to be computed once and we can obtain it for any $\de$ by either interpolating the table or computing numerically with code to be shared (runtime $< 1$ second).
For the parameter of the first mini-batch size $m^{(1)}$, a value of $50$ performs robustly in all experiments.

We provide the sample complexity below with the proof in Appx.~\ref{sec:proof_sample_normal}. Particularly, $T^*(\Delta) \rightarrow DN$ as $\Delta \rightarrow 0$, and $T^*(\Delta) = D m^{(1)}$ when $\Delta \geq 2 B_{\mathrm{Normal}}(\de/D')$ $\sqrt{(N / m^{(1)}-1)/(N-1)}$.

\begin{proposition}\label{prop:sample_normal}
Let $x^*$ be the best arm and $\Delta$ be the minimal normalized gap of means from other arms, defined as $\min_{i\neq x^*} \frac{\mu_{x^*}-\mu_i}{\sg_{x^*}+\sg_i}$ when using marginal variance estimate $\hat{\sg}_i$ and $\min_{i\neq x^*} \frac{\mu_{x^*}-\mu_i}{\sg_{x^*,i}}$ when using pairwise variance estimate $\hat{\sg}_{x,i}$. If Assump.~\ref{assum:normal} holds, with a probability at least $1-\de$ Racing-Normal draws no more rewards than
\begin{equation}
T^*(\Delta) = D\left\lceil \frac{N}{(N-1)\frac{\Delta^2}{4 B_{\mathrm{Normal}}^2(\de/D')} + 1}\right\rceil_{m^{(1)}} \label{eq:stopping_normal}
\end{equation}
where $\lceil n \rceil_{m} \defeq m 2^{\lceil \log_2 n/m \rceil} \wedge N \geq n, \forall n\leq N$. $D'\defeq D$ if using $\hat{\sg}_i$ and is $D-1$ if using $\hat{\sg}_{x,i}$.
\end{proposition}

\subsection{Variance Reduction for Random Rewards with Control Variates}
The difficulty of MABs depends heavily on the ratio of the mean gap to the reward noise, $\Delta$. To improve the signal noise ratio, we exploit the control variates technique \cite{wilson1984variance} to reduce the reward variance. Consider a variable $h_{i,n}$ whose expectation $\eE_{n\sim [N]}[h_{i,n}]$ can be computed efficiently. The residue reward $l_{i,n} - h_{i,n} + \eE_n[h_{i,n}]$ has the same mean as $l_{i,n}$ and the variance is reduced if $h_{i,n} \approx l_{i,n}$. In the Bayesian inference experiment where the factor $f_n(X=i)=p(\by_n|X=i)$, we adopt a similar approach as \citet{wang2013variance} and take the Taylor expansion of $l_{i,n}$ around a reference point $\hat{\by}$ as
\begin{equation}
l_{i,n} \approx l_i(\hat{\by}) + \bg_i^T (\by_n - \hat{\by}) + \ha (\by_n - \hat{\by})^T H_i (\by_n - \hat{\by}) \defeq h_{i,n}
\end{equation}
where $\bg_i$ and $H_i$ are the gradient and Hessian matrix of $\log p(\by|i)$ respectively evaluated at $\hat{\by}$. $\eE[h_{i,n}]$ can computed analytically with the first two moments of $\by_n$.
A typical choice of $\hat{\by}$ is $\eE[\by]$.

The control variate method is mostly useful for Racing-Normal. For algorithms depending on a reward bound $C$ in order to get a tight bound for $l_{i,n} - h_{i,n}$ it requires a more restrictive condition for C as in \citet{bardenet2015markov} and we might end up with an even more conservative strategy in general cases.

%

\section{Related Work}\label{sec:related_work}
The Gumbel-Max trick has been exploited in \citet{kuzmin2005optimum,PaYu11a,maddison14astar} for different problems. The closest work is \citet{maddison14astar} where this trick is extended to draw continuous random variables with a Gumbel process, reminiscent to adaptive rejection sampling.

Our work is closely related to the optimal arm identification problem for MABs with a fixed confidence. This is, to our knowledge, the first work to consider MABs with a finite population. The proposed algorithms tailored under this setting could be of interest beyond the discrete sampling problem. The normal assumption in Sec.~\ref{sec:racing-normal} is similar to UCB-Normal in \citet{auer2002finite} but the latter assumes a normal distribution for \textit{individual} rewards and will perform poorly when it does not hold.

The bounds in Sec.~\ref{sec:racing} are based on subsampling-based MH algorithms in \citet{bardenet2014towards,korattikara2013austerity}. The proposed algorithm extends those ideas from MH to discrete sampling. In fact, let $x$ and $x'$ be the current and proposed value in an MH iteration, Racing-EBS and Racing-Normal reduce to the algorithms in \citet{bardenet2014towards} and \citet{korattikara2013austerity} respectively if we set 
\begin{align}
&\cX=\{x,x'\},\quad f_0(1)=u~p(x)q(x'|x),\nn\\
&f_0(2)=p(x')q(x|x'),\quad f_n(x) = p(\by_n|x)
\end{align}
where $p(x)$ is the prior distribution, $u\sim\mathrm{Uniform}[0,1]$ and $q(\cdot|\cdot)$ is the proposal distribution. The difference with \citet{bardenet2014towards} is that we distribute the error $\de$ evenly across $t$ in Eq.~\ref{eq:G_ebs} while \citet{bardenet2014towards} set $\de_t = (p-1)/(p (T^{(t)})^p)\de$ with $p$ a free parameter. The differences with \citet{korattikara2013austerity} are that we take a doubling schedule for $m^{(t)}$ and replace the t-test with the normal assumption. We find that our algorithms are more efficient and robust than both original algorithms in practice. Moreover, the binary Gibbs sampling in Appx.~F of \citet{korattikara2013austerity} is also a special case of Racing-Normal with $D=2$. Therefore, Alg.~\ref{alg:racing} provides a unifying approach to a family of subsampling-based samplers.

The variance reduction technique is similar to the proxies in \citet{bardenet2015markov}, but the control variate here is a function in the data space while the proxy in the latter is a function in the parameter space. We do not assume the posterior distribution is approximate Gaussian and our algorithm works with multi-modal distributions.

It is important not the confuse the focus of our algorithm for the big $N$ problem in Eq.~\ref{eq:tilde_p} with other algorithms that address sampling for a large state space (big $D$) or similarly a high-dimensional vector of discrete variables (exponentially large $D$). The combination of these two approaches for problems with both big $N$ and big $D$ is possible but beyond the scope of this paper.

\section{Experiments}\label{sec:experiments}
Since this is the first work to discuss efficient discrete sampling for problem (\ref{eq:tilde_p}), we compare the adapted lil'UCB, Racing-EBS, Racing-Normal with the exact sampler only. We report the result of Racing-Normal in real data experiments only as the speed gains of the other two are marginal.

\vspace{-0.1cm}
\subsection{Synthetic Data}
\begin{figure*}[tb]%
\centering
  \subfigure[$p(X)$]{%
    \label{fig:toy_p}%
    \includegraphics[width=0.25\textwidth]{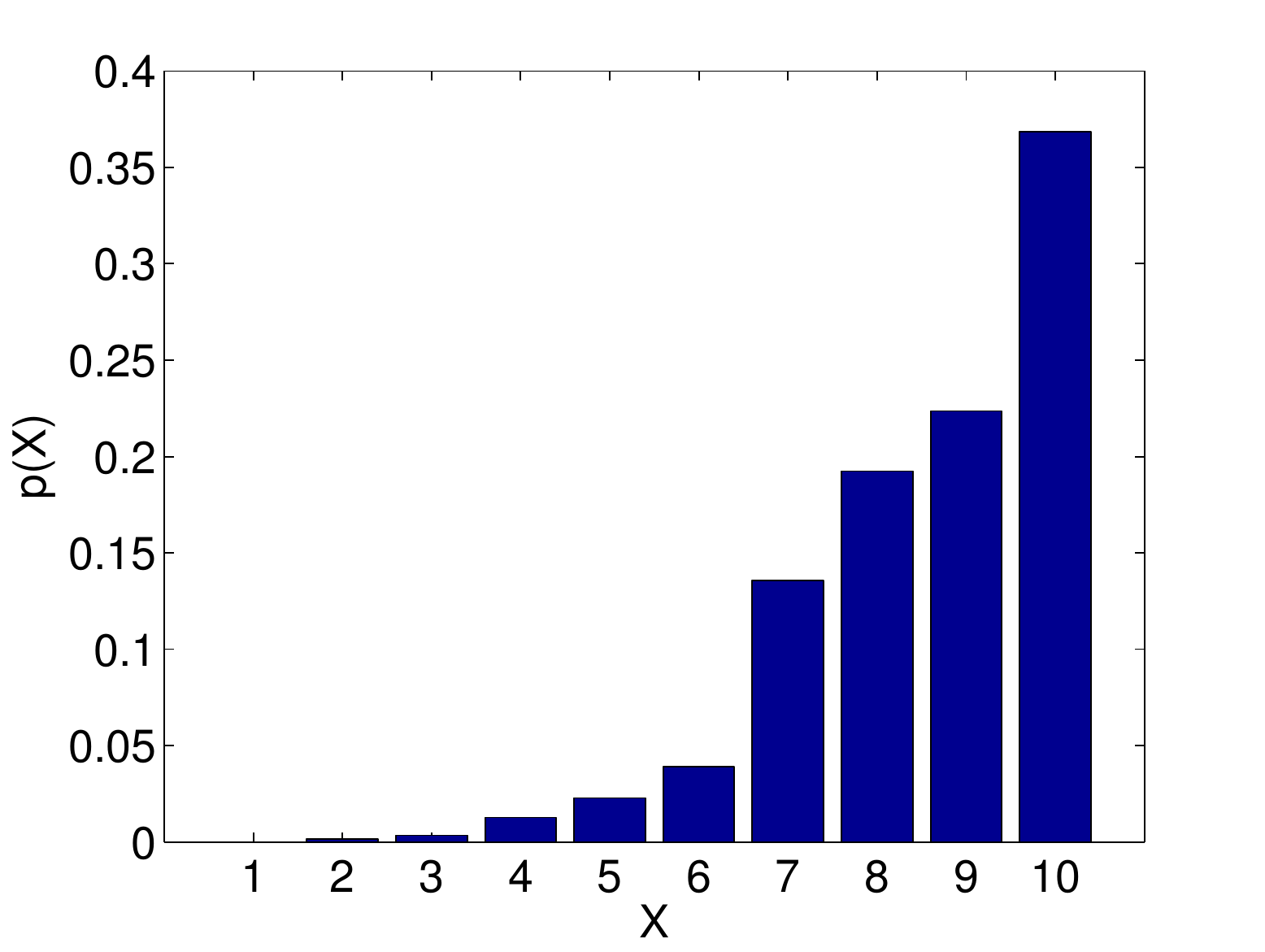}%
  }%
  ~ 
  \subfigure[$\sg=0.1$, very hard]{%
    \label{fig:toy_error_1}%
    \includegraphics[width=0.25\textwidth]{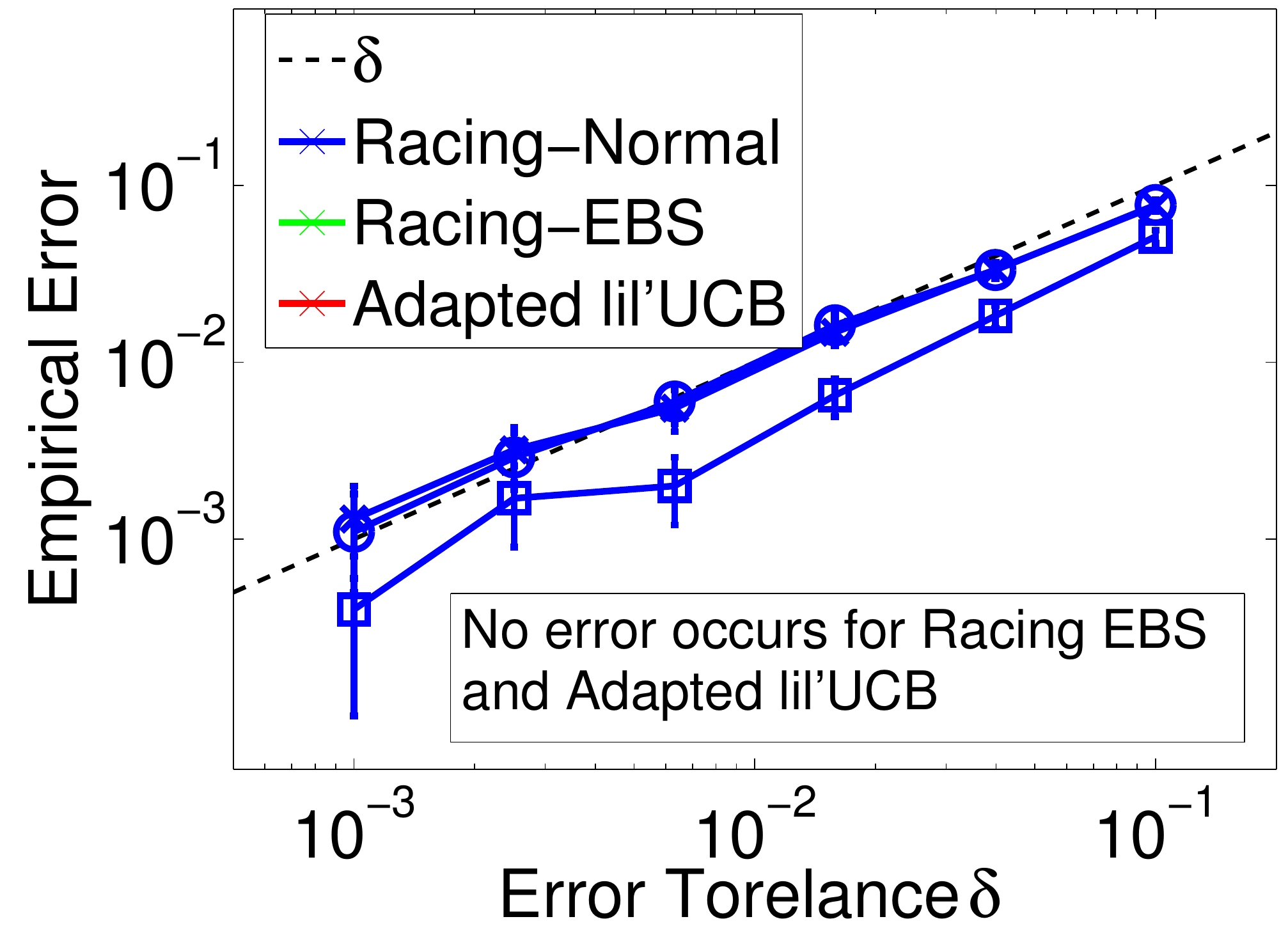}%
  }%
  ~
 \subfigure[$\sg=10^{-4}$, easy]{%
    \label{fig:toy_error_2}%
    \includegraphics[width=0.25\textwidth]{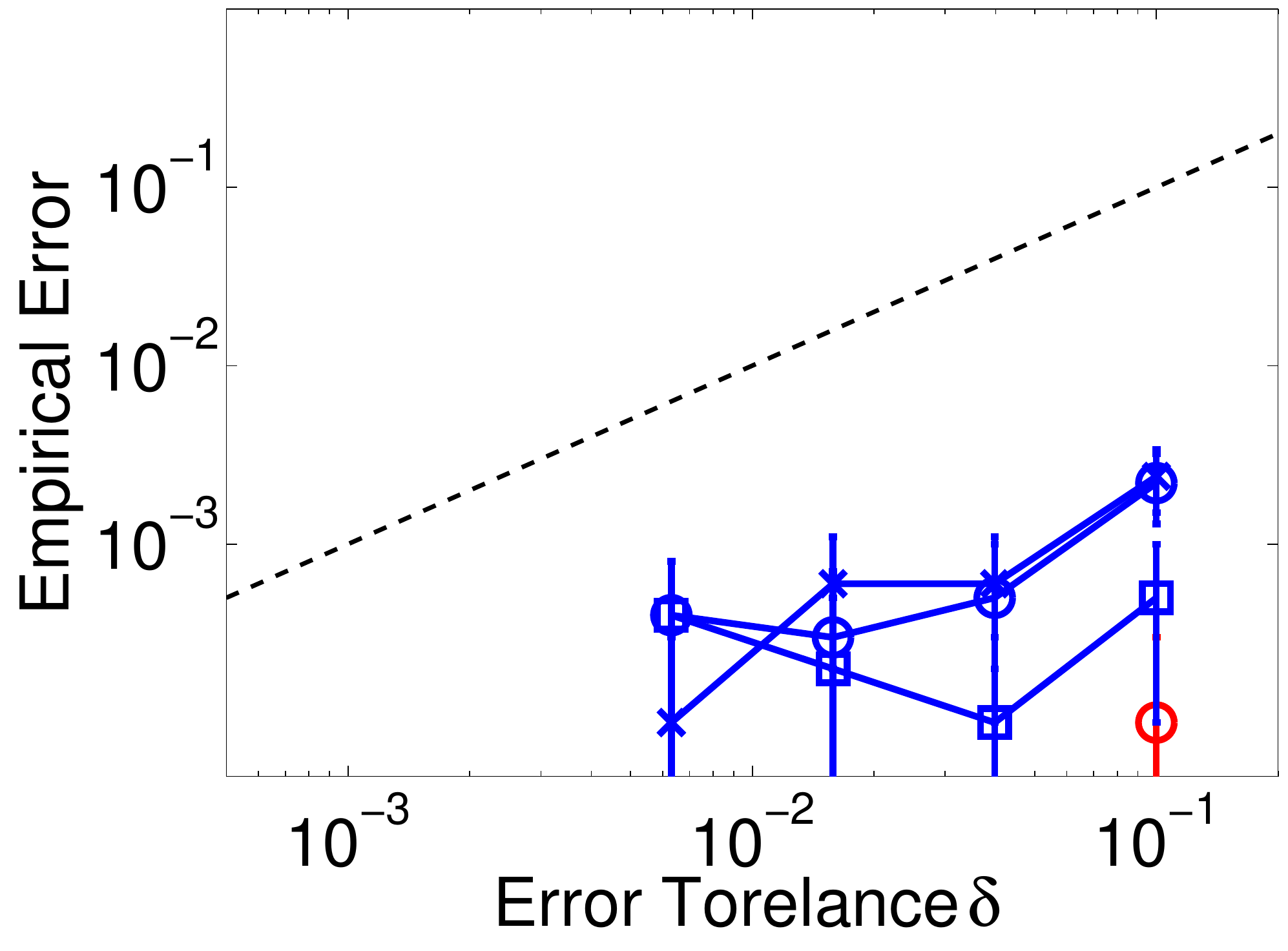}%
  }%
  ~
  \subfigure[$\sg=10^{-5}$, very easy]{%
    \label{fig:toy_error_3}%
    \includegraphics[width=0.25\textwidth]{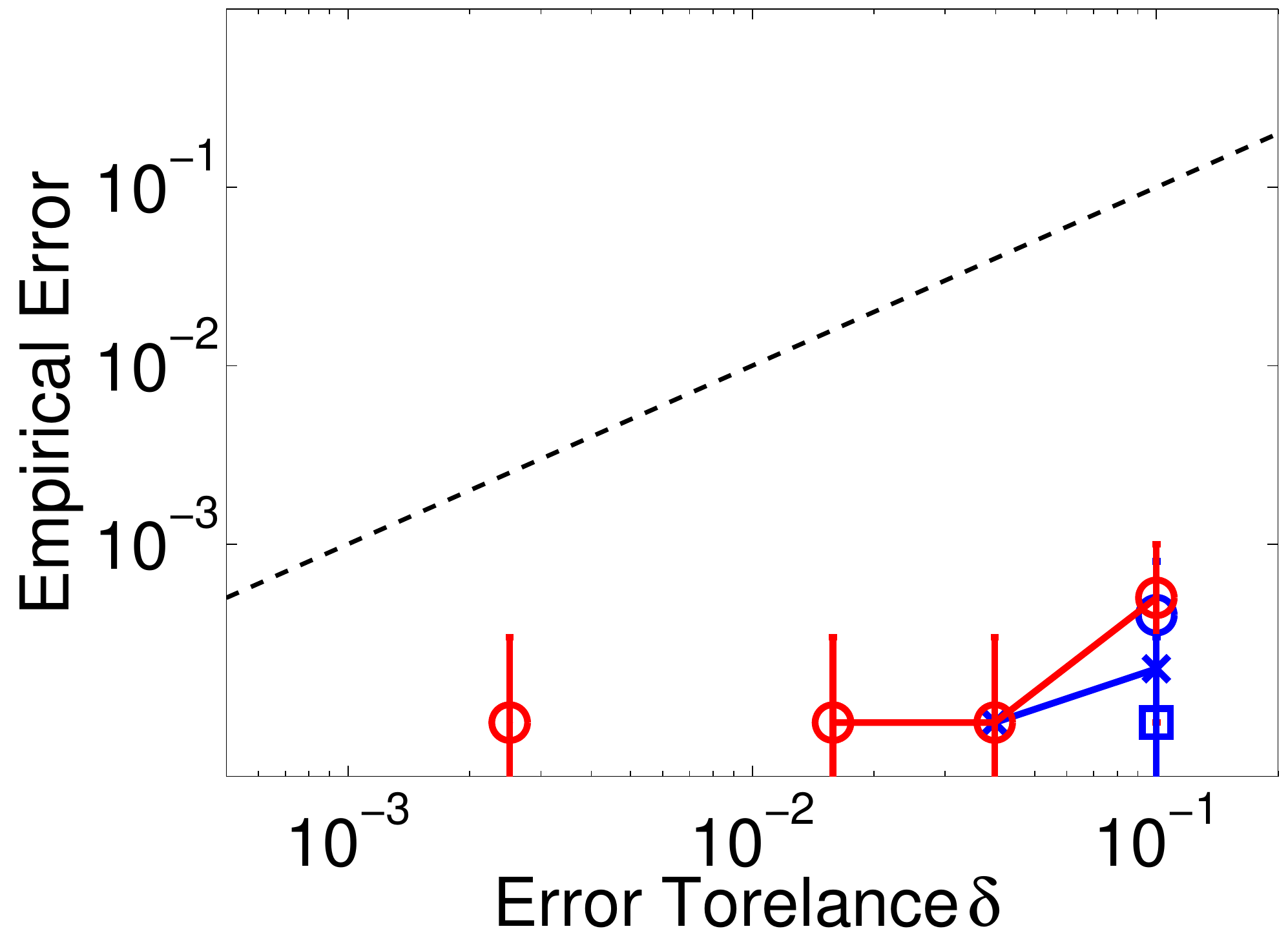}%
  }%
  \\
  \subfigure[Uncertainty bounds $G(T)$ with $\de=0.1$.]{%
    \label{fig:toy_bounds}%
    \includegraphics[width=0.25\textwidth]{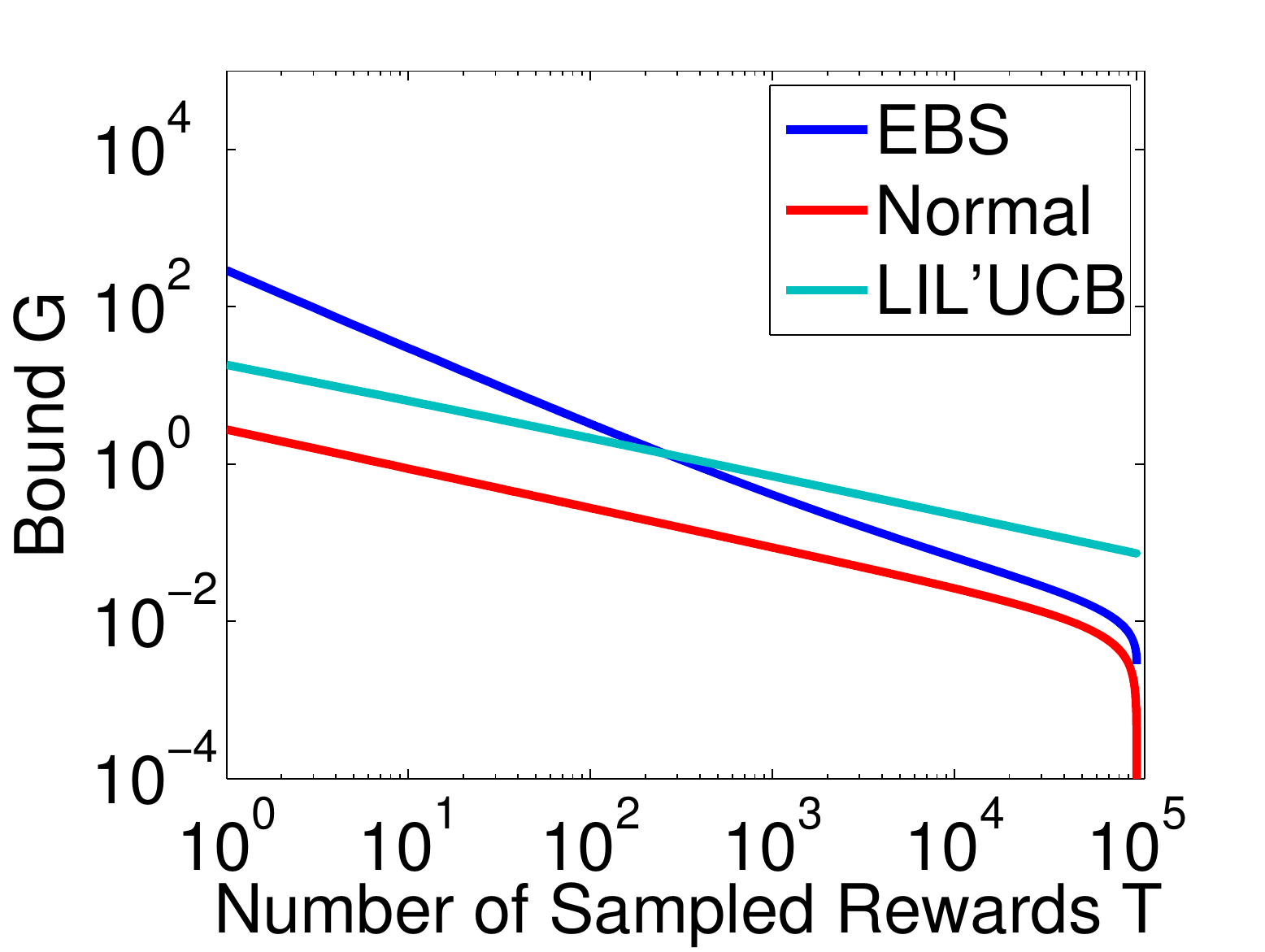}%
  }%
  ~
  \subfigure[$\sg=0.1$]{%
    \label{fig:toy_data_1}%
    \includegraphics[width=0.25\textwidth]{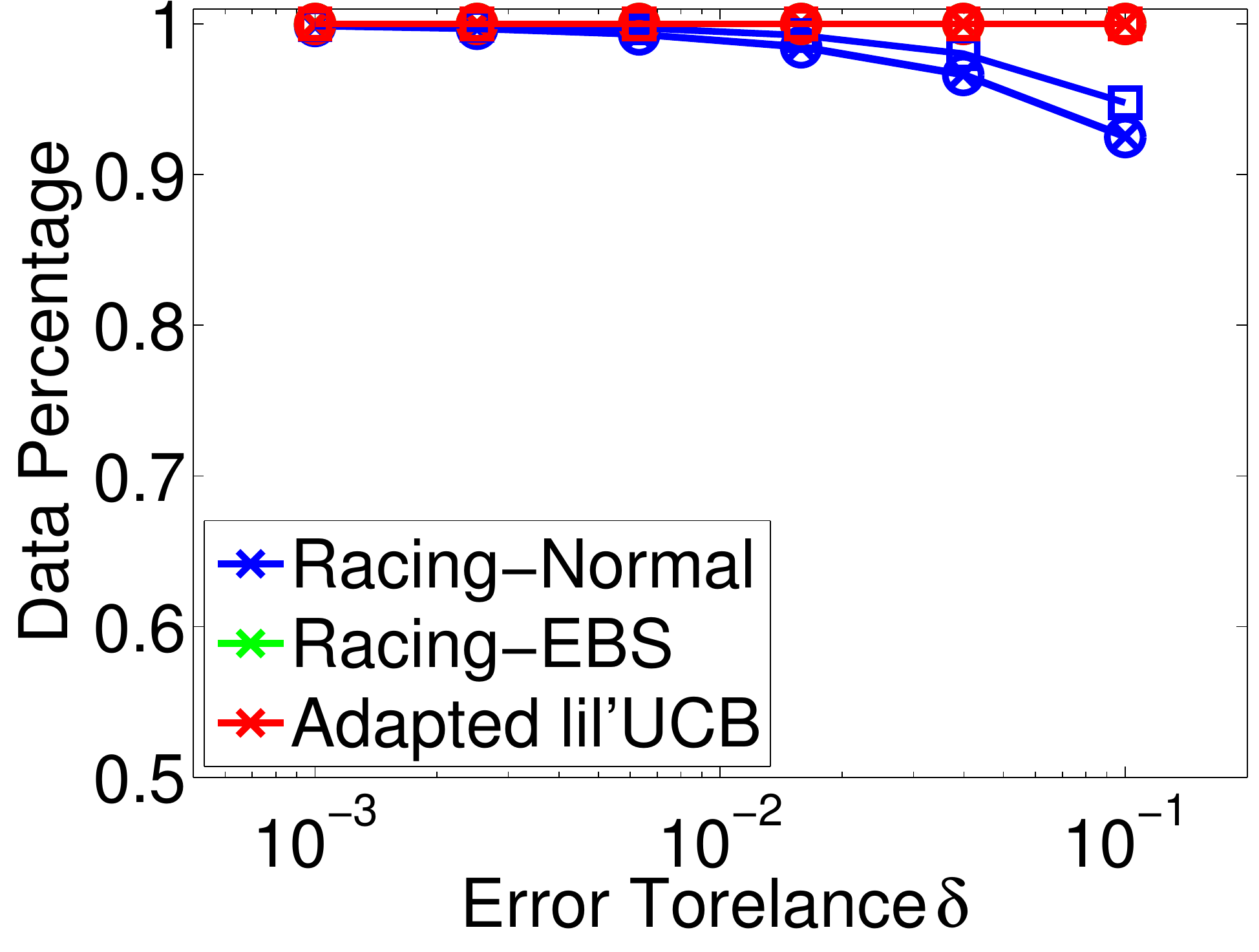}%
  }%
  ~
  \subfigure[$\sg=10^{-4}$, in log scale]{%
    \label{fig:toy_data_2}%
    \includegraphics[width=0.25\textwidth]{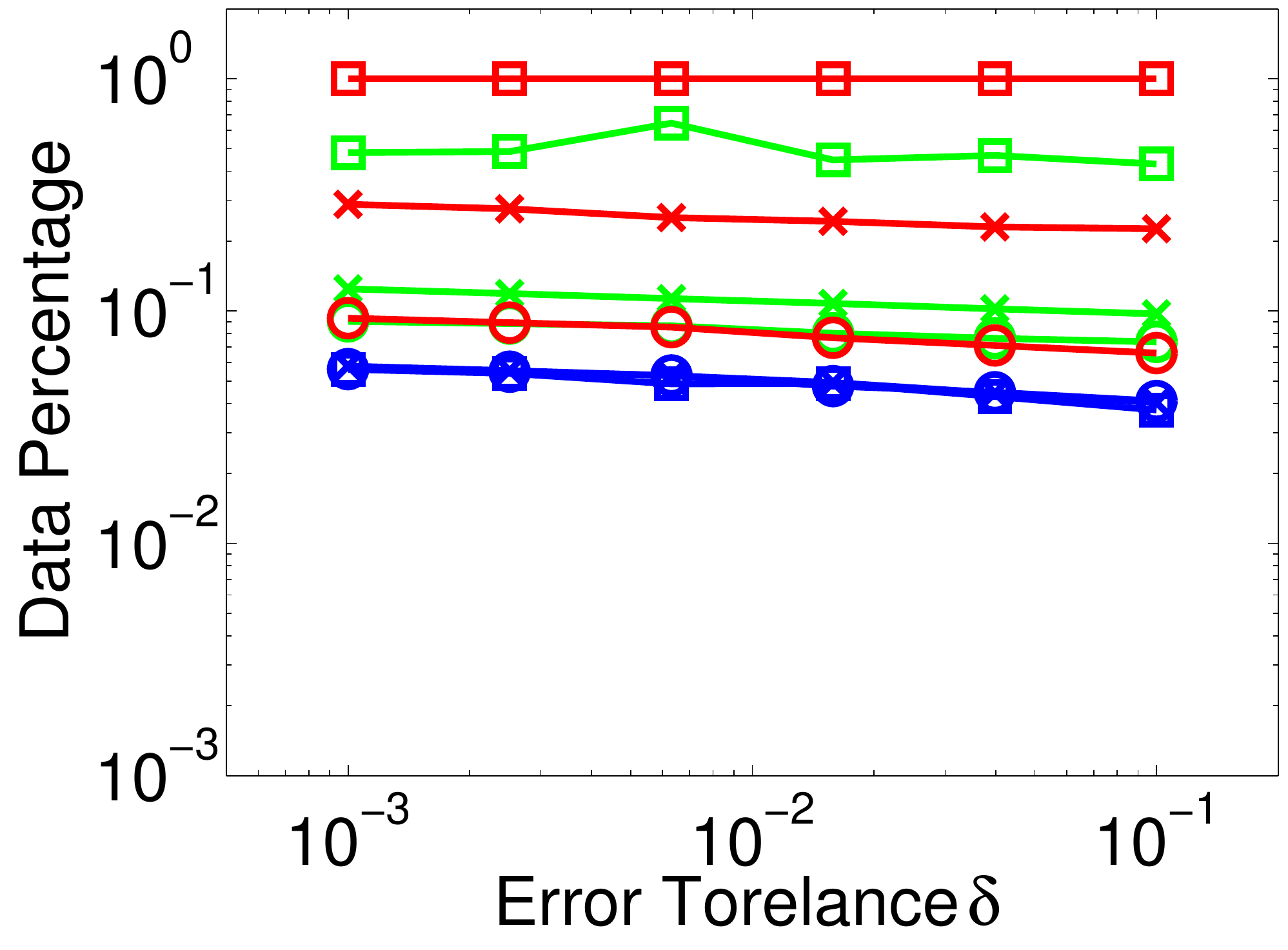}%
  }%
  ~
  \subfigure[$\sg=10^{-5}$, in log scale]{%
    \label{fig:toy_data_3}%
    \includegraphics[width=0.25\textwidth]{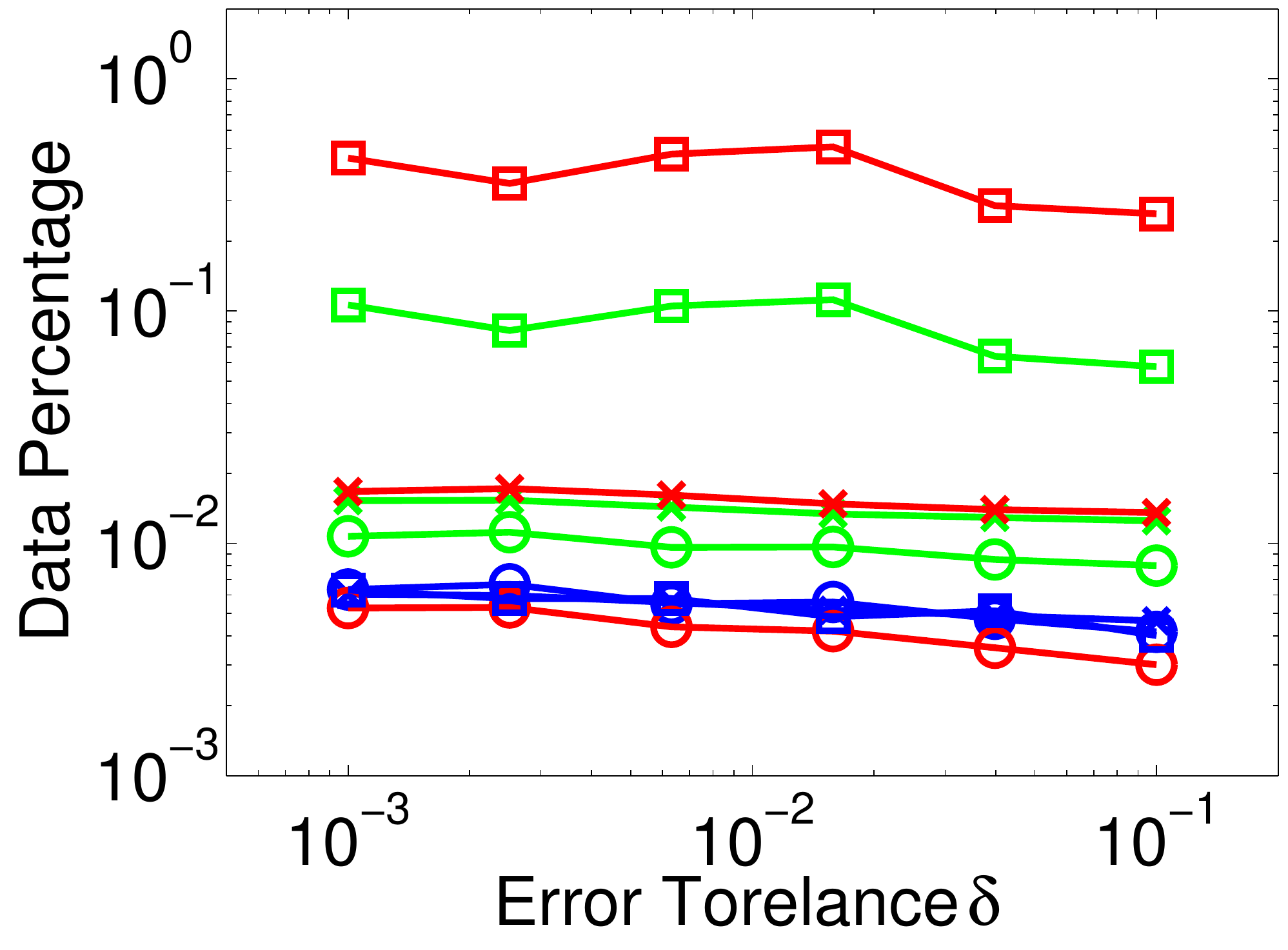}%
  }%
  \vspace{-0.3cm}
  \caption{Synthetic data. (\subref{fig:toy_error_1},\subref{fig:toy_error_2},\subref{fig:toy_error_3}) Estimated error with $95\%$ confidence interval. \textbf{Plots not shown if no error occured}. (\subref{fig:toy_data_1},\subref{fig:toy_data_2},\subref{fig:toy_data_3}) proportion of sampled rewards. $l_{i,n}$ is sampled from Normal ($\times$), Uniform ($\bigcirc$) and LogNormal ($\square$) distributions. Plots of Racing-Normal overlap in (\subref{fig:toy_data_1},\subref{fig:toy_data_2},\subref{fig:toy_data_3}).}\label{fig:toy}
\vspace{-0.3cm}
\end{figure*}

We construct a distribution with $D=10$ by sampling $N=10^5$ rewards of $l_{i,n}$ for each state from one of the three distributions $\cN(0,1)$, $\mathrm{Uniform}[0,1]$, $\mathrm{LogNormal}(0,2)$. We normalized $l_{i,n}$ to have a fixed distribution $p(X)$ in Fig.~\ref{fig:toy_p} and a reward variance $\sg^2$ that controls the difficulty.
The normal distribution is the ideal setting for Racing-Normal, and the uniform distribution is desirable for adapted lil'UCB and Racing-EBS as the reward bound is close to $\sg$. The LogNormal distribution, whose ex.\ kurtosis $\approx 4000$, is difficult for all due to the heavy tail.
We use a tight bound $C=\max\{l_{i,n} - l_{i,n'}\}$ for Racing-EBS. We set the scale parameter of adapted lil'UCB with $C/2$ and other parameters with the heuristic setting in \citet{jamieson2014lil}. Racing uses the pairwise variance estimate.

Fig.~\ref{fig:toy_error_1}-\subref{fig:toy_error_3} show the empirical error of best arm identification by drawing $10^4$ samples of $X$ for each setting and vary the target error bound $\de\in[10^{-3},0.1]$. The bound appears very loose for lil'UCB and Racing-EBS but is sharp for Racing-Normal when the noise is large (\ref{fig:toy_error_1}) and $\de\ll 1$. This is consistent with the direct comparison of uncertainty bounds in Fig.~\ref{fig:toy_bounds}.
Consequently, given the same error tolerance $\de$ Racing-Normal requires much fewer rewards than the other conservative strategies in all the settings except when $\sg=10^{-5}$ and $l_{i,n}\sim \mathrm{Uniform}[0,1]$, as shown in Fig.~\ref{fig:toy_data_1}-\subref{fig:toy_data_3}.
We verify the observations with more experiments in Appx.~\ref{sec:extra_exp_toy}  with $D\in\{2, 100\}$ and marginal estimate $\hat{\sg}_i$.



Surprisingly, Racing-Normal performs robustly regardless of reward distributions with the first mini-batch size $m^{(1)}=50$ while it was shown in \citet{bardenet2014towards} that the algorithm with the same normal assumption in \citet{korattikara2013austerity} failed with LogNormal even when $m^{(1)}=500$. The dramatic improvement in robustness is mainly due to our doubling scheme where central limit theorem applies quickly with $m^{(t)}$ increasing exponentially. We do not claim that the single trick will solve the problem completely because there still exist cases in theory with extremely heavy-tailed reward distributions where our normal assumption does not hold and the algorithm will fail to meet the confidence level. In practice, we do not observe that pathological case in any of the experiments.

\begin{figure*}[tbh]
\centering
\begin{tabular}{l|r}
\begin{minipage}[t]{0.64\textwidth}%
  \centering
  \subfigure[Stock index return $r_t$]{%
    \label{fig:arch_return}%
    \includegraphics[width=0.48\textwidth]{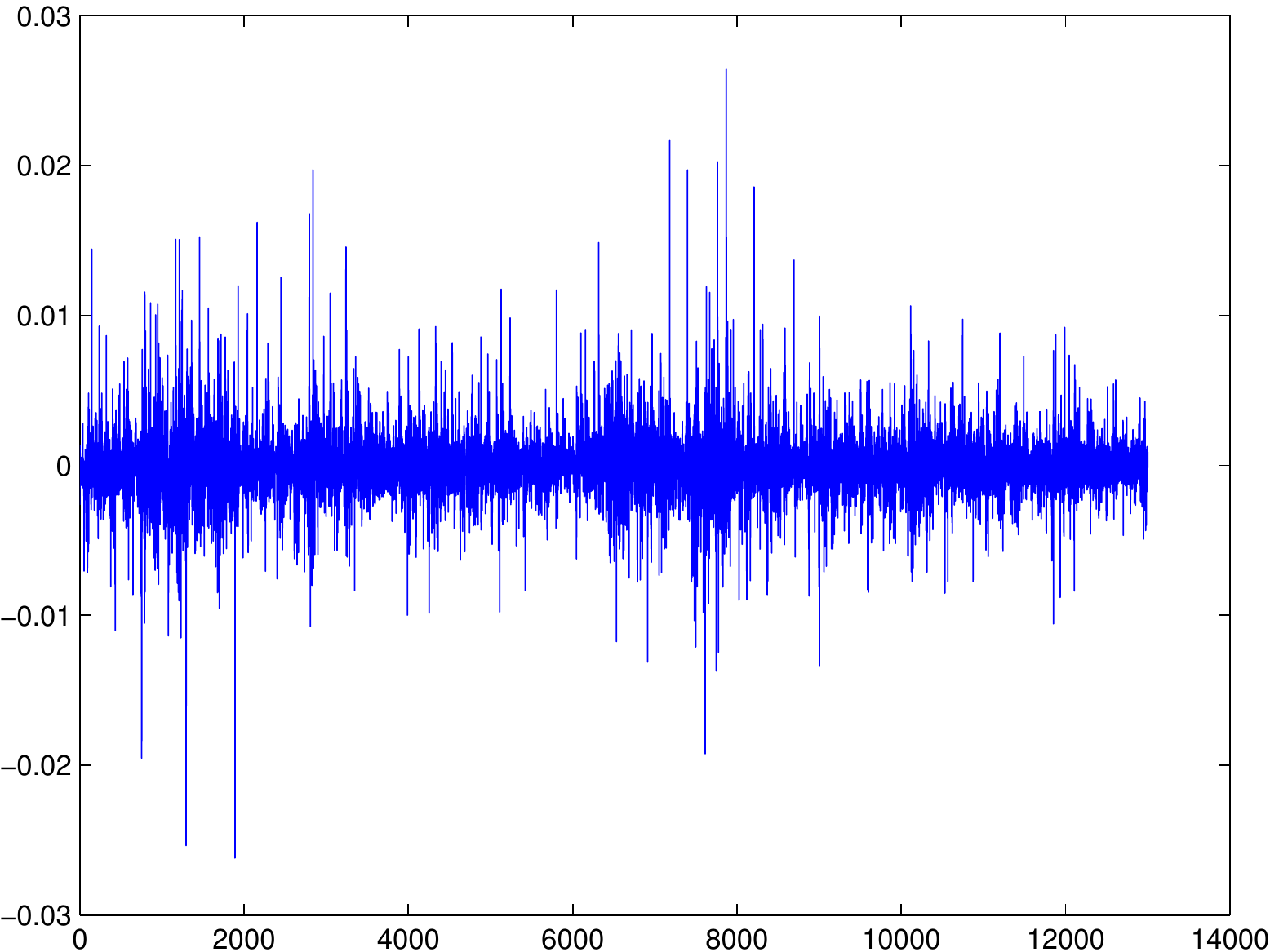}%
  }%
  ~
  \subfigure[Estimated $\log p(q|\br)$]{%
    \label{fig:arch_log_pq}%
    \includegraphics[width=0.5\textwidth]{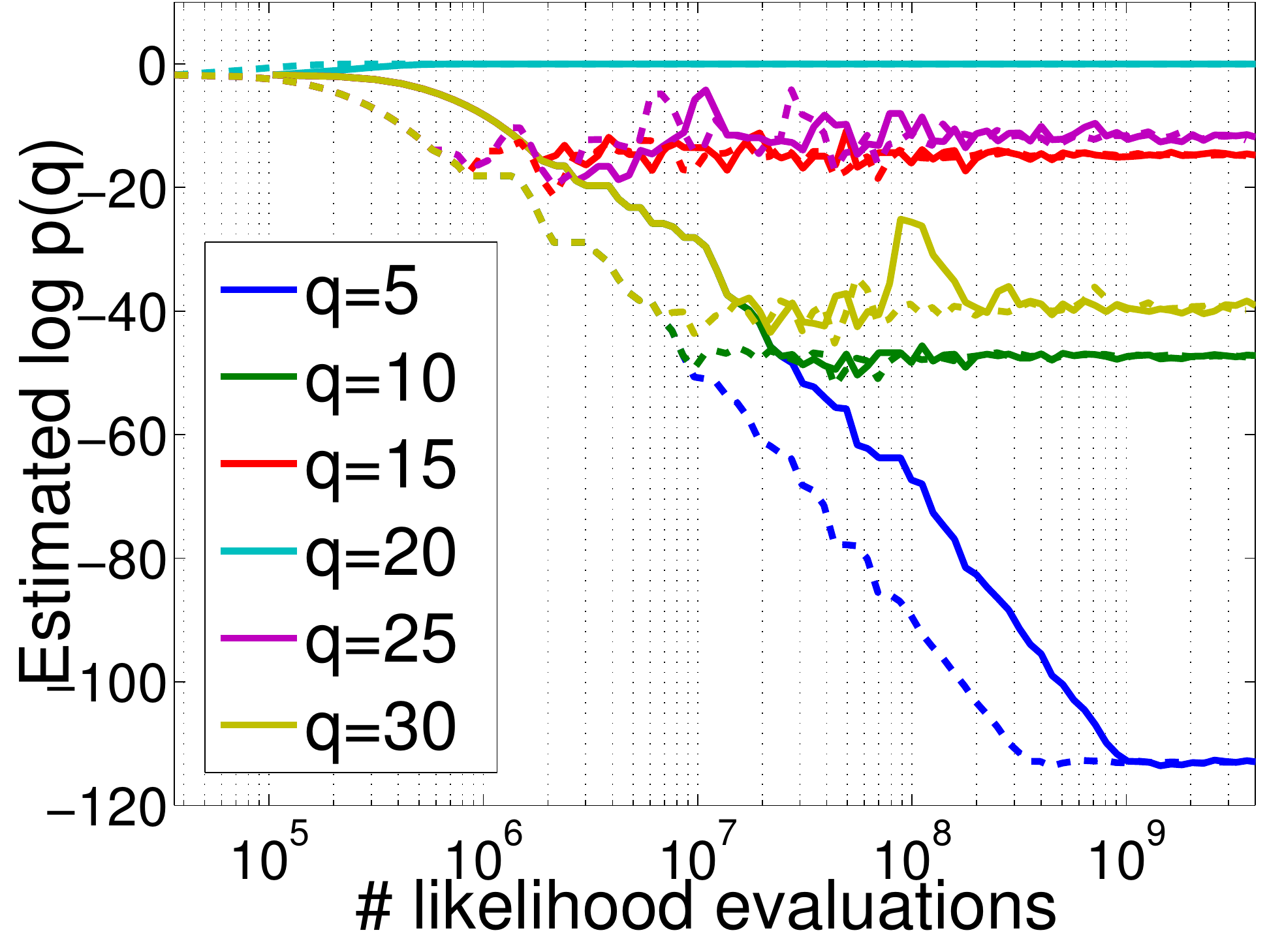}%
  }%
  \\
  \vspace{-0.2cm}
  \subfigure[Adjusted post. $\tilde{p}(q|\br)$]{%
    \label{fig:arch_adjusted_pq}%
    \includegraphics[width=0.5\textwidth]{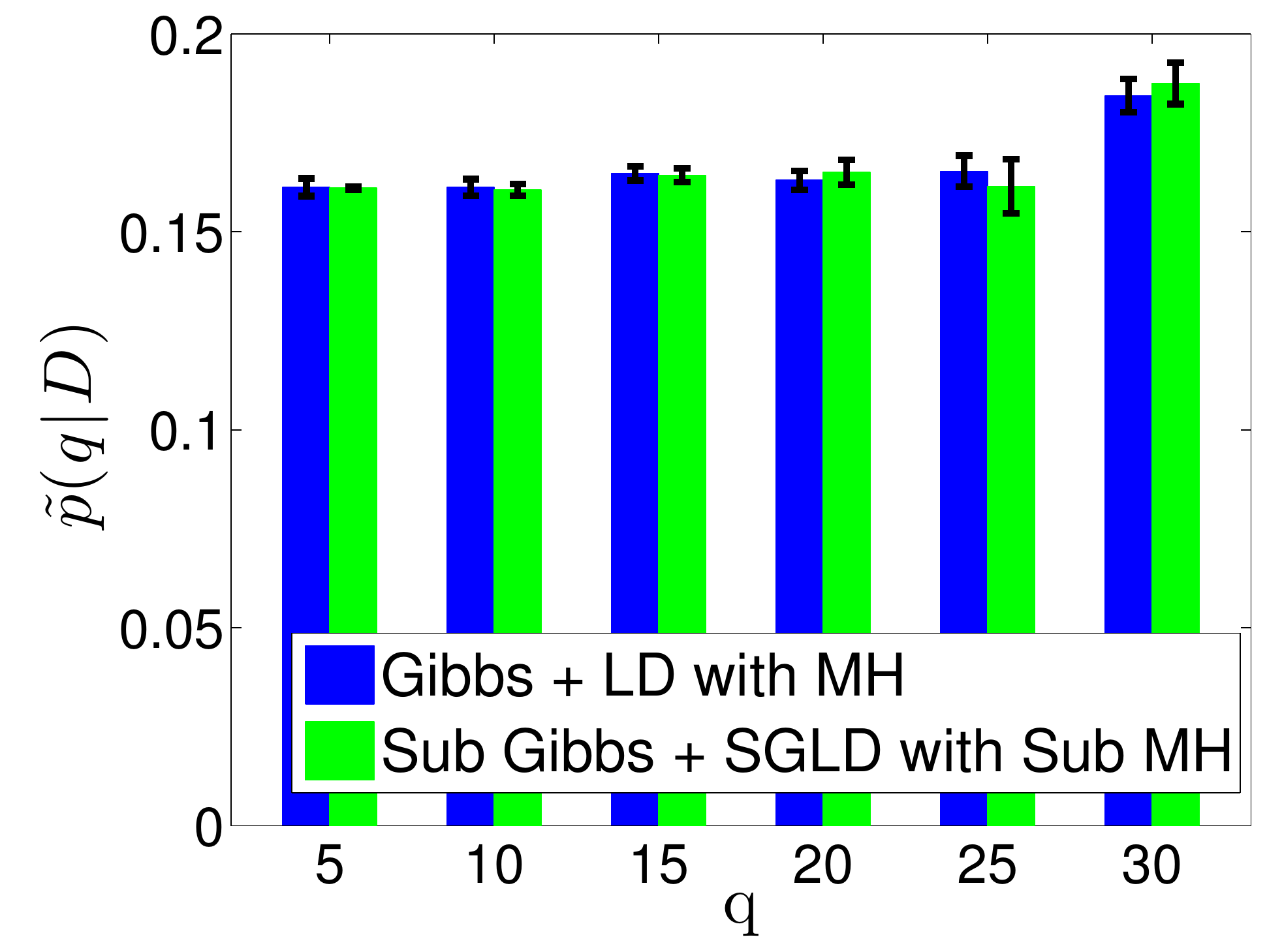}%
  }%
  ~
  \subfigure[Auto-correlation of $q$]{%
    \label{fig:arch_xcov_s}%
    \includegraphics[width=0.5\textwidth]{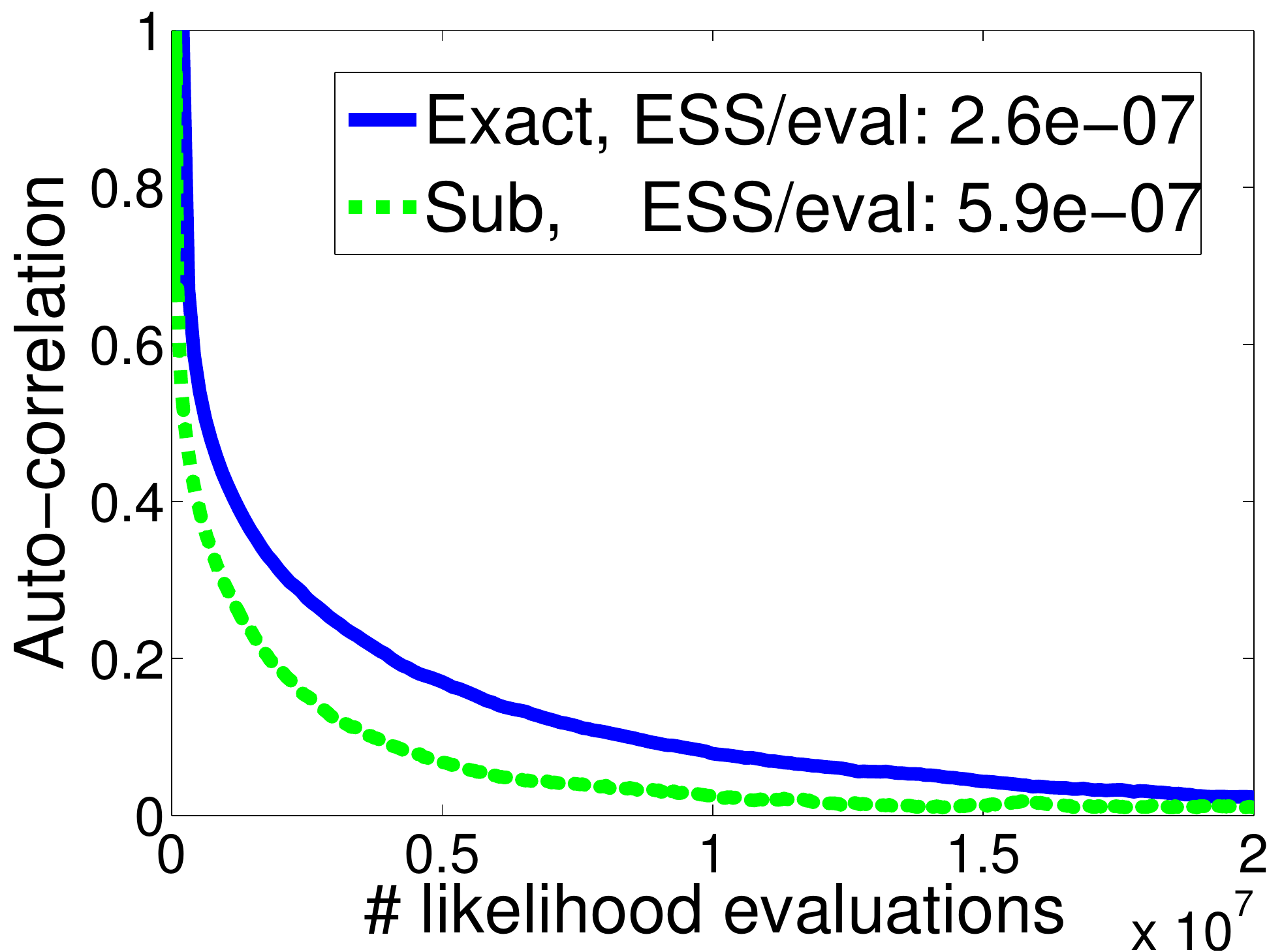}%
  }%
  \vspace{-0.2cm}
  \caption{Bayesian ARCH Model Selection. Solid: exact, dashed: approximate using Sub Gibbs + SGLD with Sub MH.}%
\end{minipage}%
&
\begin{minipage}[t]{0.32\textwidth}%
  \centering
  \subfigure[$B^3$ F-1 vs \#factor evaluations]{%
    \label{fig:rexa_f1}%
    \includegraphics[width=\textwidth]{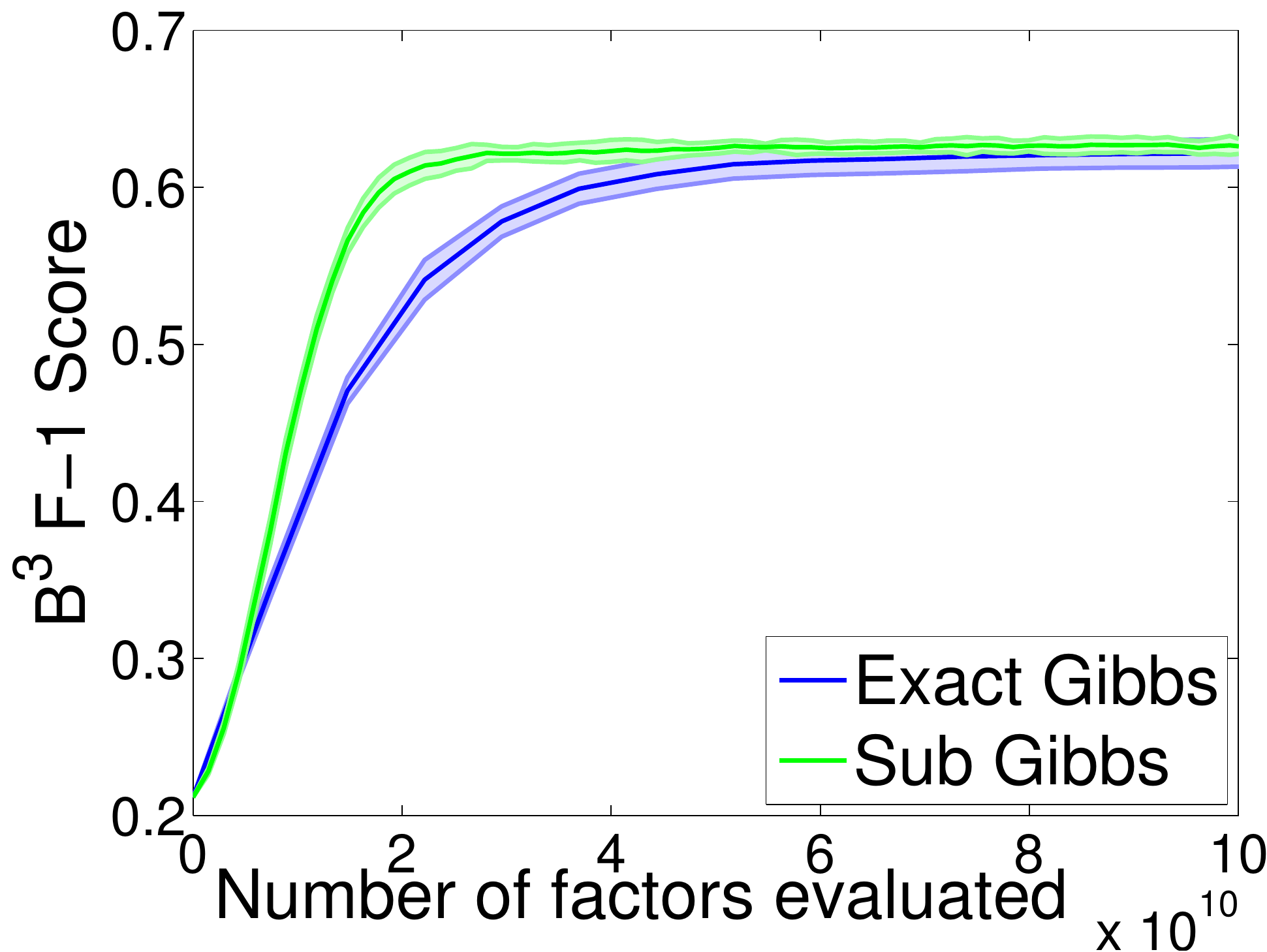}%
  }%
  \\
  \vspace{-0.2cm}
  \subfigure[$B^3$ F-1 vs iteration]{%
    \label{fig:rexa_f1_iter_speed}%
    \includegraphics[width=\textwidth]{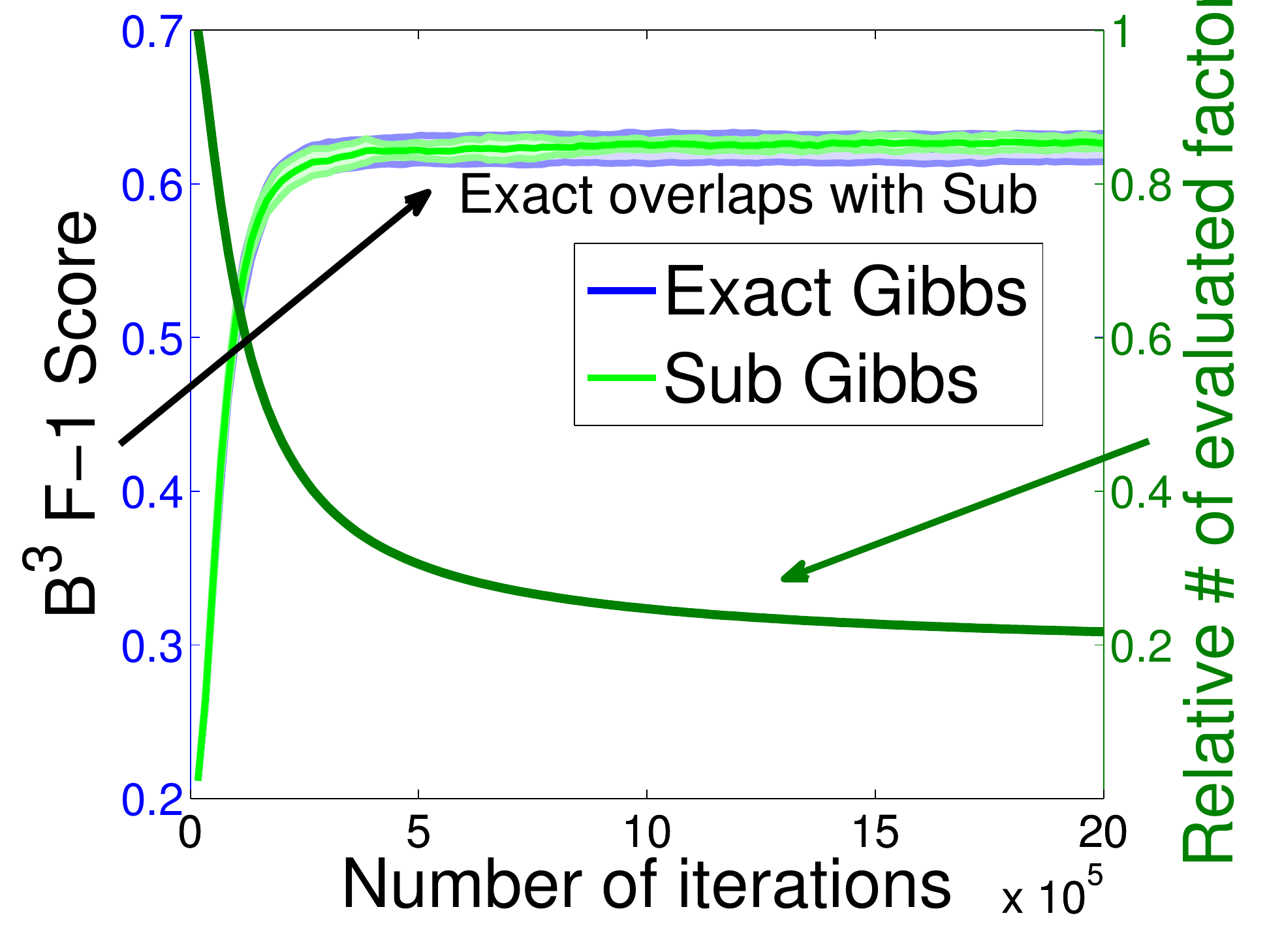}%
  }%
  \vspace{-0.2cm}
  \caption{Author Coreference. Bigger $B^3$ F-1 score is better.}%
\end{minipage}%
\end{tabular}
\vspace{-0.2cm}
\end{figure*}


\vspace{-0.1cm}
\subsection{Bayesian ARCH Model Selection}
We evaluate Racing-Normal in a Bayesian model selection problem for the auto-regressive conditional heteroskedasticity (ARCH) models. The discrete sampler is integrated in the Markov chain as a building component to sample the hierarchical model. 
Specifically, we consider a mixture of ARCHs for the return $r_t$ of stock price series with student-t innovations, each component with a different order $q$:
\begin{align}
&r_t = \sg_t z_t,~z_t\overset{iid}\sim t_{\nu}(0,1),\quad \sg_t^2 = \al_0 + \sum_{i=1}^q \al_i r_{t-i}^2,\nn\\
&q \sim \mathrm{Discrete}(\bpi),\quad \alpha_i,\nu \overset{iid}\sim \mathrm{Gamma}(1,1) \nn
\end{align}
where $\bpi=\{\pi_q: q\in\mathbb{Q}\}$ is the prior distribution of a candidate model in the set $\mathbb{Q}$.
The random variables to infer include the discrete model choice $q$ and continuous parameters $\{\alpha_i\}_{i=0}^q, \nu$. 
We adopt the augmented MCMC algorithm in \citet{CarlinChib95} to avoid transdimensional moves. 
We apply subsampling-based scalable algorithms to sample all variables with subsampled observations $\{r_t\}$:
Racing-Normal Gibbs for $q$, stochastic gradient Langevin dynamics (SGLD) \cite{welling2011bayesian} corrected with Racing-Normal MH (Sec.~\ref{sec:related_work}) for $\alpha_i$ and $\nu$. We use adjusted priors $\tilde{\pi}_q$ as suggested by \citet{CarlinChib95} for sufficient mixing between all models and tune them with adaptive MCMC. The adjusted posterior $\tilde{p}(q|\br)\propto \tilde{\pi}_qp(\br|q)$ is then close to uniform and the value $\pi_q/\tilde{\pi}_q$ provides an estimate to the real unnormalized posterior $p(q|\br)$. Control variates are also applied to reduce variance. Details of the sampling algorithm are provided in Appx.~\ref{sec:extra_exp_arch}.

We apply the model on the 5-minute Shanghai stock exchange composite index of one year consisting of about 13,000 data points (Fig.~\ref{fig:arch_return}). $\mathbb{Q}=\{5,10,15,20,25,30\}$. We set $m^{(1)}=50$ and $\de=0.05$. The control variate method reduces the reward variance by 2$\sim$3 orders of magnitude. Fig.~\ref{fig:arch_log_pq} shows the estimated log-posterior of $q$ by normalizing $\pi_q/\tilde{\pi}_q$ in the adaptive MCMC as a function of the number of likelihood evaluations (proportional to runtime). The subsampling-based sampler (Sub) converges about three times faster. We then fix $\tilde{\pi}_q$ for a fixed stationary distribution and run MCMC for $10^5$ iterations to compare Sub with the exact sampler. The empirical error rates for Racing-Normal Gibbs and MH are about $4\times 10^{-4}$ and $2\times 10^{-3}$ respectively. Fig.~\ref{fig:arch_adjusted_pq} shows estimated adjusted posterior with 5 runs, and \subref{fig:arch_xcov_s} compares the auto-correlation of sample $q$. Sub obtains over twice the effective sample size without noticeable bias after the burn-in period.

\vspace{-0.1cm}
\subsection{Author Coreference}
We then study the performance in a large-scale graphical model inference problem. The author coreference problem for a database of scientific paper citations is to cluster the mentions of authors into real persons. \citet{singh2012monte} addressed this problem with a conditional random field model with pairwise factors. The joint and conditional distributions are respectively
\begin{align}
p_{\bta}(\by|\bx) &\propto \exp\left(\sum_{y_i=y_j, i\neq j, \forall i,j } f_{\bta}(x_i, x_j)\right),\nn\\
p_{\bta}(Y_i=y_i|\by_{-i},\bx) &\propto \exp\left(\sum_{y_j\in C_y = \{j: y_j=y, j\neq i\}} f_{\bta}(x_i, x_j)\right) \nn
\end{align}
where $\bx=\{x_i\}_{i=1}^N$ is the set of observed author mentions and $y_i\in\mathbb{N}^+$ is the cluster index for $i$'th mention. The factor $f_{\bta}(x_i, x_j)$ measures the similarity between two mentions based on author names, coauthors, paper title, etc, parameterized by $\bta$. In the conditional distribution, $y_i$ can take a value of any non-empty cluster or another empty cluster index.
When a cluster $C_y$ contains a lot of mentions, a typical case for common author names, the number of factors to be evaluated $N_y=|C_y|$ will be large.
We consider the MAP inference problem with fixed $\bta$ using annealed Gibbs sampling \cite{finkel2005incorporating}.
We apply Racing-Normal to sample $Y_i$ by subsampling $C_y$ for each candidate value $y$. An important difference of this problem from Eq.~\ref{eq:tilde_p} is that $N_y \neq N_{y'}, \forall y\neq y'$ and $N_y$ has a heavy tail distribution. We let the mini-batch size depend on $N_y$ with details provided in Appx.~\ref{sec:extra_exp_rexa}.

We run the experiment on the union of an unlabeled DBLP dataset of BibTex entries with about 5M authors and a Rexa corpus of about 11K author mentions with 3160 entries labeled.
We monitor the clustering performance on the labeled subset with the $B^3$ F-1 score \cite{bagga1998algorithms}. We use $\de=0.05$ and the empirical error rate is about $0.046$. The number of candidate values $D$ varies in $2\sim 215$ and $N_y$ varies in $1\sim 1829$ upon convergence. Fig.~\ref{fig:rexa_f1} shows the F-1 score as a function of the number of factor evaluations with 7 random runs for each algorithm. Sub Gibbs converges about three times faster than exact Gibbs. Fig.~\ref{fig:rexa_f1_iter_speed} shows F-1 as a function of iterations that renders almost identical behavior for both algorithms, which suggests negligible bias in Sub Gibbs. The relative number of the evaluated factors of sub to exact Gibbs indicates about a 5-time speed up near convergence. The initial speed up is small because every cluster is initialized with a single mention, i.e.\ $N_y=1$.

\vspace{-.1cm}
\section{Discussion}\label{sec:discussion}
We consider the discrete sampling problem with a high degree of dependency and proposed three approximate algorithms under the framework of MABs with theoretical guarantees. The Racing algorithm provides a unifying approaches to various subsampling-based Monte Carlo algorithms and also improves the robustness of the original MH algorithm in \citet{korattikara2013austerity}. This is also the first work to discuss MABs under the setting of a finite reward population.

Empirical evaluations show that Racing-Normal achieves a robust and the highest speed-up among all competitors. Whilst adaptive lil'UCB shows inferior empirical performance to Racing-Normal, it has a better sample complexity w.r.t.\ the number of arms $D$. It will be a future direction to combine the bound of Racing-Normal with other MAB algorithms including lil'UCB for a better scalability in $D$.
Another important problem is on how to relax the assumptions for Racing-Normal without sacrificing the performance.

It would also be an interesting direction to extend our work to draw continuous random variables efficiently with the Gumbel process \cite{maddison14astar}. In continuous state space, there are infinitely many ``arms" and a naive application of our algorithm will lead to infinitely large error bound. This problem can be alleviated with algorithms for contextual MAB problems.

\section*{Acknowledgements} 

We thank Matt Hoffman for helpful discussions on the connection of our work
to the MAB problems. We also thank all the reviewers for their constructive comments.
We acknowledge funding from the Alan Turing Institute, Google, Microsoft Research
and EPSRC Grant EP/I036575/1.



\bibliography{disc_austerity}
\bibliographystyle{icml2016}

\newpage
\appendix

\section{Proofs}
\subsection{Proof of Prop.~\ref{prop:bound}}\label{sec:proof_bound}
\begin{proof}
For a discrete state space, the total variation is equivalent to half of $L_1$ distance between two probability vectors. Denote by $\hat{p}(X=i|\beps)$ the distribution of the output of the approximate algorithm conditioned on the vector of Gumbel variables $\beps$, and $x(\beps)$ the solution of Eq.~\ref{eq:disc_gumbel} as a function of $\beps$. According to the premise of Prop.~\ref{prop:bound}, $\hat{p}(X=x(\beps)|\beps)\geq 1-\de, \forall \beps$. We can bound the $L_1$ error of the conditional probability as
\begin{align}
&\sum_{i\in \cX} \left|\hat{p}(X=i|\beps) - \de_{i,x(\beps)}\right| \nn\\
&= \left|\hat{p}(X=x(\beps)|\beps) - 1\right| + \sum_{i\neq x(\beps)} \left|\hat{p}(X=i|\beps)\right| \leq 2\de, \forall \beps
\end{align}
where $\de_{i,j}$ is the Kronecker delta function. Then we can show
\begin{align}
&\|\hat{p}(X) - p(X)\|_{\mathrm{TV}} \nn\\
&= \ha \sum_{i\in \cX} \left|\tilde{p}(X=i) - p(X=i)\right| \nn\\
&= \ha \sum_{i\in \cX} \left| \int_{\beps} \left(\hat{p}(X=i|\beps) - \de_{i,x(\beps)}\right) \mathrm{d} P(\beps)\right| \nn\\
&\leq \ha \sum_{i\in \cX} \int_{\beps} \left|\hat{p}(X=i|\beps) - \de_{i,x(\beps)}\right| \mathrm{d} P(\beps) \nn\\
&= \ha \int_{\beps} \left(\sum_{i\in \cX} \left|\hat{p}(X=i|\beps) - \de_{i,x(\beps)}\right|\right) \mathrm{d} P(\beps) \nn\\
&\leq \de
\end{align}
\end{proof}

\subsection{Sketch of the proof of Prop.~\ref{prop:lil'ucb}}\label{sec:proof_lilucb}
\begin{proof}
As the proof of this proposition is almost identical to the proof of \citet{jamieson2014lil}, we only outlines the difference due to the adaptation. In the proof of Thm.~2 in \citet{jamieson2014lil}, the i.i.d.\ assumption for rewards from each arm was used only in Lemma 3 to provide Chernoff's bound and Hoeffding's bound. As noted in Sec.~6 of \citet{hoeffding1963probability} those bounds would still hold when rewards are sampled from a finite population without replacement. Therefore, when $T^{(t)}<N$ all the bounds hold for adapted lil'UCB.

When $T_i^{(t)}=N$, the second modification sets the upper bound of the mean estimate to $\hat{\mu}^{(t)}$. That is a valid upper bound of $\mu_i$, in fact much tighter than the bound in the original algorithm because $\hat{\mu}^{(t)}_i=\mu_i$ exactly when the entire population is observed.

Therefore, as long as $T_i^{(t)} \leq N, \forall i$, Theorem 2 in \citet{jamieson2014lil} applies to adapted lil'UCB with modification 1 and 2 only.

With the third modification, $T^{(t)}$ could never be bigger than $N$ at the stopping time, which proves the second part of Prop \ref{prop:lil'ucb}. The proof can then be concluded if we can show modification 3 does not change the output of adapted lil'UCB with the first two modifications only. This is true because if we do not stop when the selected arm $i$ satisfies $T_i^{(t)}=N$, we do not need to update the upper bound of $i$ because the estimated mean is already exact. Since no upper bound is changed, the arm $i$ will always be chosen for now on and eventually the original stopping criterion of $T_i^{(t)} \geq 1 + \lambda \sum_{j\neq i}T_j(t)$ is met and the same arm $i$ will be returned.
\end{proof}

\subsection{Proof of Prop.~\ref{prop:racing}}\label{sec:proof_racing}
\begin{proof}
Denote by $x^{(t)}$ the arm with the highest estimated mean at iteration $t$ and $x^*$ the optimal arm with the highest true mean, $\mu_{x^*}>\mu_i, \forall i\neq x^*$. If Alg.~\ref{alg:racing} does not stop in the first $t^*-1$ iterations, the estimated means of all the survived arms become exact at the last iteration $t^*$, $\hat{\mu}_i^{(t^*)} = \mu_i$ because we require $T^{(t^*)}=N$. Then $x^{(t^*)}=x^*$. As we require $G(\de,T=N,\hat{\sg},C)=0, \forall \de, \hat{\sg}, C$, all the sub-optimal arms will be eliminated by the last iteration and the algorithm always returns the correct best arm. This proves the upper bound of the sample size of $ND$.

Now to prove the confidence level, all we need to show is that with at least a probability $1-\de$ arm $x^*$ survived all the iterations $t < t^*$.

Let us first consider the case when Alg.~\ref{alg:racing} uses the marginal variance estimate $\hat{\sg}_i^{(t)}$. Let the events
\begin{align}
&E_i = \left\{\exists t < t^*, \hat{\mu}_i^{(t)} - \mu_i > G\left(\frac{\de}{D}, T^{(t)}, \hat{\sg}_i^{(t)}, C_i\right)\right\}, \forall i\neq x^* \nn\\
&E_{x^*} = \left\{\exists t < t^*, -\hat{\mu}_{x^*}^{(t)} - (-\mu_{x^*}) > G\left(\frac{\de}{D}, T^{(t)}, \hat{\sg}_i^{(t)}, C_i\right)\right\} \label{eq:E_1}
\end{align}
Applying condition Eq.~\ref{eq:condition_G} and the union bound, we get
$
P(\cup_{i\in\cX} E_i) \leq \sum_{i\in\cX} E_i = \de.
$
So with a probability at least $1-\de$, none of those events will happen. In that case for any iteration $t < t^*$,
\begin{align}
&\hat{\mu}_x - \hat{\mu}_{x^*} = (\hat{\mu}_x - \mu_x) - (\hat{\mu}_{x^*} - \mu_{x^*}) + (\mu_x - \mu_{x^*}) \nn\\
&< G\left(\frac{\de}{D}, T^{(t)}, \hat{\sg}_x^{(t)}, C_x\right) + G\left(\frac{\de}{D}, T^{(t)}, \hat{\sg}_{x^*}^{(t)}, C_{x^*}\right)
\end{align}
So arm $x^*$ won't be eliminated at iteration $t$.

Similarly, for the case when Alg.~\ref{alg:racing} uses the pairwise variance estimate $\hat{\sg}_{x,i}^{(t)}$, let the events
\begin{align}
E_{i,x} = \bigg\{&\exists t <t^*, (\hat{\mu}_i^{(t)}-\hat{\mu}_{x^*}^{(t)}) - (\mu_i-\mu_{x^*}) \nn\\
&> G\left(\frac{\de}{D-1}, T^{(t)}, \hat{\sg}_i^{(t)}, C_{i}+C_{x^*}\right)\bigg\}, \forall i\neq x^* \label{eq:E_2}
\end{align}
Applying condition Eq.~\ref{eq:condition_G} and the union bound, we get
$
P(\cup_{i\in\cX\backslash\{x^*\}} E_{i,x}) \leq \sum_{i\in\cX\backslash\{x^*\}} E_{i,x} = \de.
$
So with a probability at least $1-\de$ for any iteration $t < t^*$,
\begin{align}
\hat{\mu}_x - \hat{\mu}_{x^*} &= (\hat{\mu}_x - \hat{\mu}_{x^*}) - (\mu_x - \mu_{x^*}) + (\mu_x - \mu_{x^*}) \nn\\
&< G\left(\frac{\de}{D-1}, T^{(t)}, \hat{\sg}_{x,x^*}^{(t)}, C_{x}+C_{x^*}\right)
\end{align}
Therefore arm $x^*$ won't be eliminated at iteration $t$.
\end{proof}

\subsection{Proof of Prop.~\ref{prop:sample_normal}}\label{sec:proof_sample_normal}
\begin{proof}
Denote by $x^{(t)}$ the arm with the highest estimated mean at iteration $t$. First consider the case when Alg.~\ref{alg:racing} uses the marginal variance estimate $\hat{\sg}_i^{(t)}$.
With the condition in Eq.~\ref{eq:condition_G}, it follows that $P(\cup_{i\in\cX} E_i) \leq \sum_{i\in\cX} P(E_i) \leq \de$ where $E_i$ is defined in Eq.~\ref{eq:E_1}. So with a probability at least $1-\de$,
\begin{align}
\hat{\mu}_{x^*}^{(t)} - \hat{\mu}_i^{(t)} > & \mu_{x^*} - \mu_i - G\left(\frac{\de}{D}, T^{(t)}, \hat{\sg}_{x^*}^{(t)}, C_{x^*}\right) \nn\\
& - G\left(\frac{\de}{D}, T^{(t)}, \hat{\sg}_i^{(t)}, C_i\right), \forall i \neq x^*
\end{align}
Alg.~\ref{alg:racing} will stop by iteration $t$ if the RHS of the equation above satisfies the stopping criterion for all $i\neq x^*$, that is,
\begin{align}
\mu_{x^*} - \mu_i > 2 \Bigg(&G\left(\frac{\de}{D}, T^{(t)}, \hat{\sg}_{x^*}^{(t)}, C_{x^*}\right)\nn\\
&+ G\left(\frac{\de}{D}, T^{(t)}, \hat{\sg}_i^{(t)}, C_i\right)\Bigg), \forall i \neq x^*
\end{align}
Plugging in the definition of $G_{\mathrm{Normal}}$ in Eq.~\ref{eq:G_normal} and applying the assumption $\hat{\sg}_i^{(t)}=\sg_i$, we will get 
\begin{equation}
\frac{\mu_{x^*} - \mu_i}{(\sg_{x^*} + \sg_{i})} > \frac{2}{\sqrt{T^{(t)}}} \left(1-\frac{T^{(t)}-1}{N-1}\right)^{1/2} B_{\mathrm{Normal}}, \forall i \neq x^*
\end{equation}
Solve the above inequality for $T^{(t)}$ and use the definition of the gap $\Delta$ we get
\begin{equation}
T^{(t)} > \frac{N}{(N-1)\frac{\Delta^2}{4 B_{\mathrm{Normal}}^2(\de/D)} + 1} \defeq \tilde{T}
\end{equation}
Since we use a doubling schedule $T^{(t)}=2T^{(t-1)}$ with $T^{(1)}=m^{(1)}$ and $T^{(t^*)}=N$, Alg.~\ref{alg:racing} stops at an iteration no later than
\begin{equation}
t = \lceil \log_2 (\tilde{T} / m^{(0)}) \rceil + 1
\end{equation}
And the total number of samples drawn by $t$ is upper bounded by $D (m^{(0)}2^{t-1} \wedge N) = T^*(\Delta)$.

Now consider the case when Alg.~\ref{alg:racing} uses the pairwise variance estimate $\hat{\sg}_{x,i}^{(t)}$.
With the condition in Eq.~\ref{eq:condition_G}, it follows with the union bound that $P(\cup_{i\in\cX\backslash\{x^*\}} E_i) \leq \sum_{i\in\cX\backslash\{x^*\}} P(E_i) \leq \de$ where $E_i$ is defined in Eq.~\ref{eq:E_2}. So with a probability at least $1-\de$,
\begin{align}
&\hat{\mu}_{x^*}^{(t)} - \hat{\mu}_i^{(t)} \nn\\
&> \mu_{x^*} - \mu_i - G\left(\frac{\de}{D-1}, T^{(t)}, \hat{\sg}_{x^*,i}^{(t)}, C_{x^*}+C_{i}\right), \forall i \neq x^*
\end{align}
Now we can follow a similar argument as in the case with marginal variance estimate and prove the proposition.
\end{proof}

\section{Table and Figure of $B_{\mathrm{Normal}}(\de,\pi_{T^{(1)}})$}\label{sec:table_B}
\begin{figure*}[tbh]
\centering
  \centering
  \includegraphics[width=0.7\textwidth]{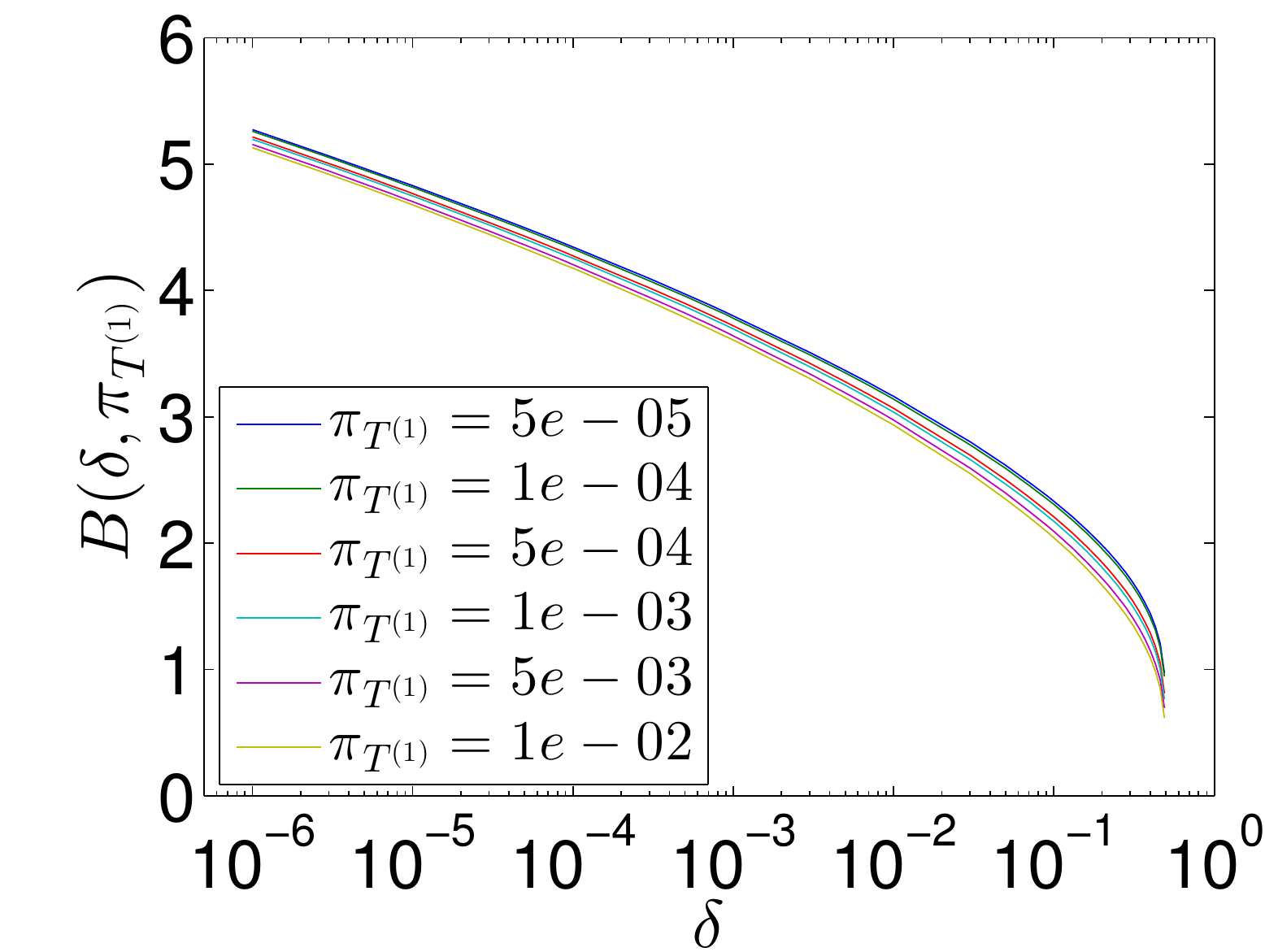}
  \caption{$B_{\mathrm{Normal}}(\de,\pi_{T^{(1)}})$}
  \label{fig:b_normal}
\end{figure*}

Table \ref{tab:B_normal} shows $B_{\mathrm{Normal}}(\de,\pi_{T^{(1)}})$ with $\de$ varying in $[10^{-6}, 0.49]$, and the proportion of the first mini-batch $\pi_{T^{(1)}}=m^{(1)}/N\in\{5\times 10^{-5}, 10^{-4},5\times 10^{-4},10^{-3},5\times 10^{-3},10^{-2}\}$. $\Phi(B)$ can be interpreted as the marginal confidence level for one iteration. The function is also shown in Fig.~\ref{fig:b_normal} for visualization. We will release the code to generate the table and to compute $B_{\mathrm{Normal}}(\de,\pi_{T^{(1)}})$ numerically.

\begin{table*}[p]
\centering
\caption{$B_{\mathrm{Normal}}(\de,\pi_{T^{(1)}})$}
\label{tab:B_normal}
\begin{tabular}{c|*{6}{c}}
& \multicolumn{6}{c}{$\pi_{T^{(1)}}$} \\
\hline
$\de$ & $5\times 10^{-5}$ & $10^{-4}$ & $5\times 10^{-4}$ & $10^{-3}$ & $5\times 10^{-3}$ & $10^{-2}$ \\
\hline
1.0e-06 & 5.27250 & 5.25978 & 5.21523 & 5.19704 & 5.15638 &  5.12982 \\
3.0e-06 & 5.06504 & 5.05294 & 5.00570 & 4.98839 & 4.94490 &  4.91964 \\
5.0e-06 & 4.96669 & 4.95260 & 4.90571 & 4.88735 & 4.84311 &  4.81818 \\
7.0e-06 & 4.89969 & 4.88715 & 4.83793 & 4.82079 & 4.77535 &  4.75037 \\
9.0e-06 & 4.85078 & 4.83613 & 4.78840 & 4.76941 & 4.72447 &  4.69877 \\
1.0e-05 & 4.82952 & 4.81667 & 4.76734 & 4.74894 & 4.70377 &  4.67696 \\
3.0e-05 & 4.60397 & 4.58943 & 4.53827 & 4.51911 & 4.47119 &  4.44485 \\
5.0e-05 & 4.49660 & 4.48108 & 4.42961 & 4.40734 & 4.36137 &  4.33158 \\
7.0e-05 & 4.42331 & 4.40694 & 4.35512 & 4.33353 & 4.28573 &  4.25692 \\
9.0e-05 & 4.36853 & 4.35265 & 4.29963 & 4.27682 & 4.22961 &  4.19891 \\
1.0e-04 & 4.34343 & 4.32914 & 4.27380 & 4.25455 & 4.20386 &  4.17608 \\
3.0e-04 & 4.09380 & 4.07655 & 4.02027 & 3.99632 & 3.94601 &  3.91438 \\
5.0e-04 & 3.97189 & 3.95539 & 3.89641 & 3.87263 & 3.82038 &  3.78605 \\
7.0e-04 & 3.88945 & 3.87195 & 3.81223 & 3.78698 & 3.73467 &  3.70026 \\
9.0e-04 & 3.82665 & 3.80955 & 3.74833 & 3.72365 & 3.66977 &  3.63422 \\
1.0e-03 & 3.79932 & 3.78066 & 3.72003 & 3.69596 & 3.64066 &  3.60812 \\
3.0e-03 & 3.51044 & 3.49128 & 3.42498 & 3.39721 & 3.34023 &  3.30253 \\
5.0e-03 & 3.36685 & 3.34814 & 3.27812 & 3.25096 & 3.19048 &  3.15168 \\
7.0e-03 & 3.26922 & 3.24913 & 3.17763 & 3.14844 & 3.08769 &  3.04691 \\
9.0e-03 & 3.19383 & 3.17396 & 3.10034 & 3.07142 & 3.00871 &  2.96758 \\
1.0e-02 & 3.16117 & 3.13913 & 3.06612 & 3.03755 & 2.97349 &  2.93484 \\
3.0e-02 & 2.80261 & 2.77885 & 2.69625 & 2.66350 & 2.59450 &  2.55058 \\
5.0e-02 & 2.61646 & 2.59217 & 2.50369 & 2.46819 & 2.39672 &  2.34862 \\
7.0e-02 & 2.48285 & 2.45761 & 2.36449 & 2.33100 & 2.25369 &  2.20744 \\
9.0e-02 & 2.37768 & 2.35127 & 2.25533 & 2.22026 & 2.14145 &  2.09317 \\
1.0e-01 & 2.33161 & 2.30704 & 2.20851 & 2.17274 & 2.09292 &  2.04351 \\
1.3e-01 & 2.21073 & 2.18499 & 2.08270 & 2.04536 & 1.96346 &  1.91214 \\
1.6e-01 & 2.10639 & 2.08030 & 1.97430 & 1.93665 & 1.85177 &  1.80027 \\
1.9e-01 & 2.01355 & 1.98592 & 1.87702 & 1.83878 & 1.75267 &  1.69949 \\
2.2e-01 & 1.92898 & 1.90035 & 1.78969 & 1.74854 & 1.66259 &  1.60660 \\
2.5e-01 & 1.84734 & 1.81893 & 1.70515 & 1.66472 & 1.57552 &  1.52056 \\
2.8e-01 & 1.76920 & 1.73957 & 1.62421 & 1.58220 & 1.49310 &  1.43584 \\
3.1e-01 & 1.69110 & 1.66145 & 1.54360 & 1.50171 & 1.41066 &  1.35354 \\
3.4e-01 & 1.61302 & 1.58274 & 1.46319 & 1.42011 & 1.32819 &  1.27094 \\
3.7e-01 & 1.52953 & 1.49919 & 1.37749 & 1.33482 & 1.24221 &  1.18303 \\
4.0e-01 & 1.44411 & 1.41048 & 1.28960 & 1.24393 & 1.15002 &  1.09455 \\
4.3e-01 & 1.33819 & 1.30896 & 1.18163 & 1.14025 & 1.04396 &  0.98381 \\
4.6e-01 & 1.20662 & 1.17447 & 1.05191 & 1.00383 & 0.91939 &  0.85273 \\
4.9e-01 & 0.97014 & 0.94399 & 0.81030 & 0.76485 & 0.69587 &  0.61783
\end{tabular}
\end{table*}

\section{Experiment Detailed Setting and Extra Results}
\subsection{More Results of the Synthetic Data Experiment}\label{sec:extra_exp_toy}
\begin{figure*}[tbh]
  \centering
  \subfigure[$\sg=0.1$, very hard]{%
    \label{fig:toy_ms_error_1}%
    \includegraphics[width=0.32\textwidth]{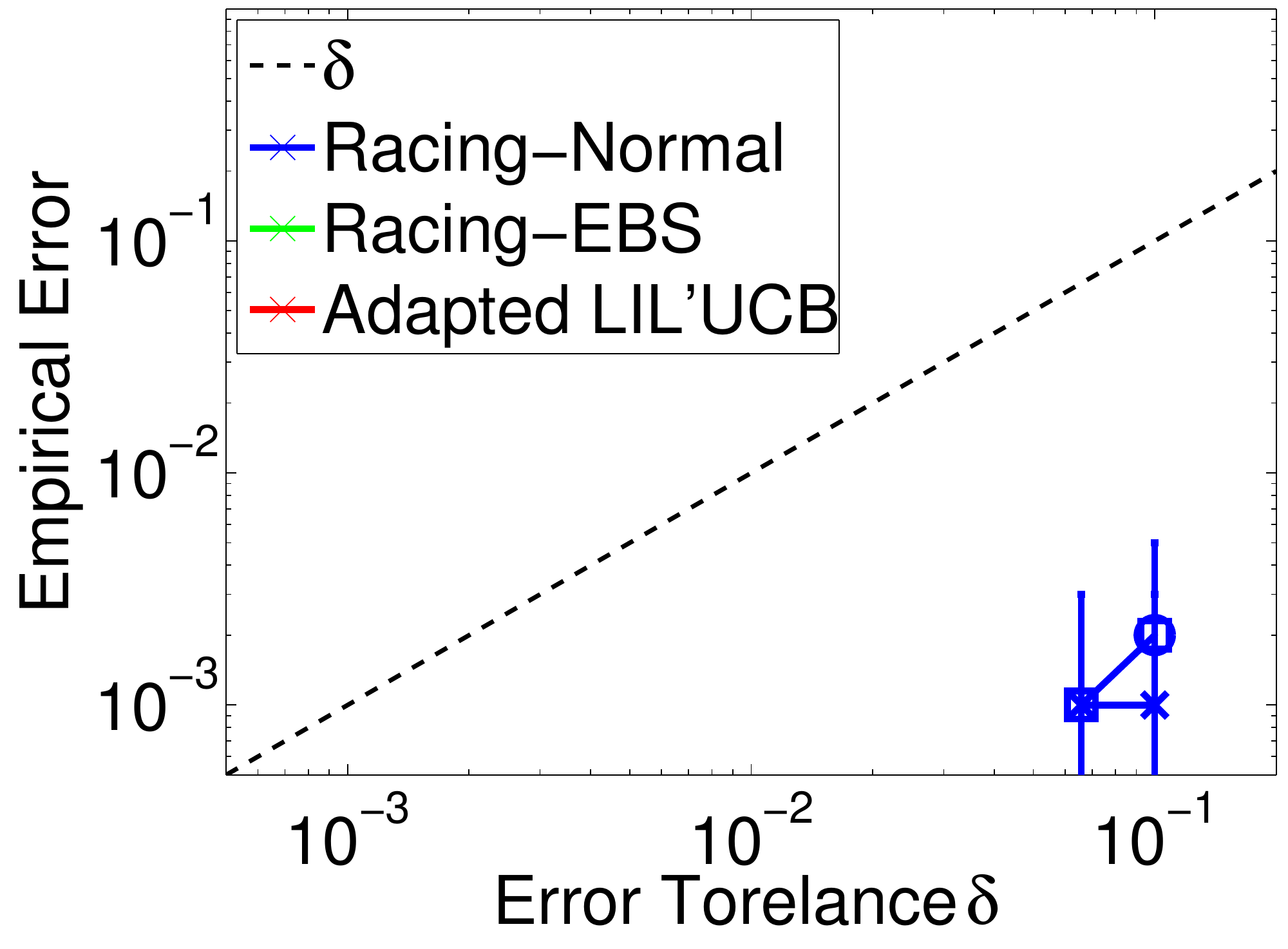}%
  }%
  ~
 \subfigure[$\sg=10^{-4}$, easy]{%
    \label{fig:toy_ms_error_2}%
    \includegraphics[width=0.32\textwidth]{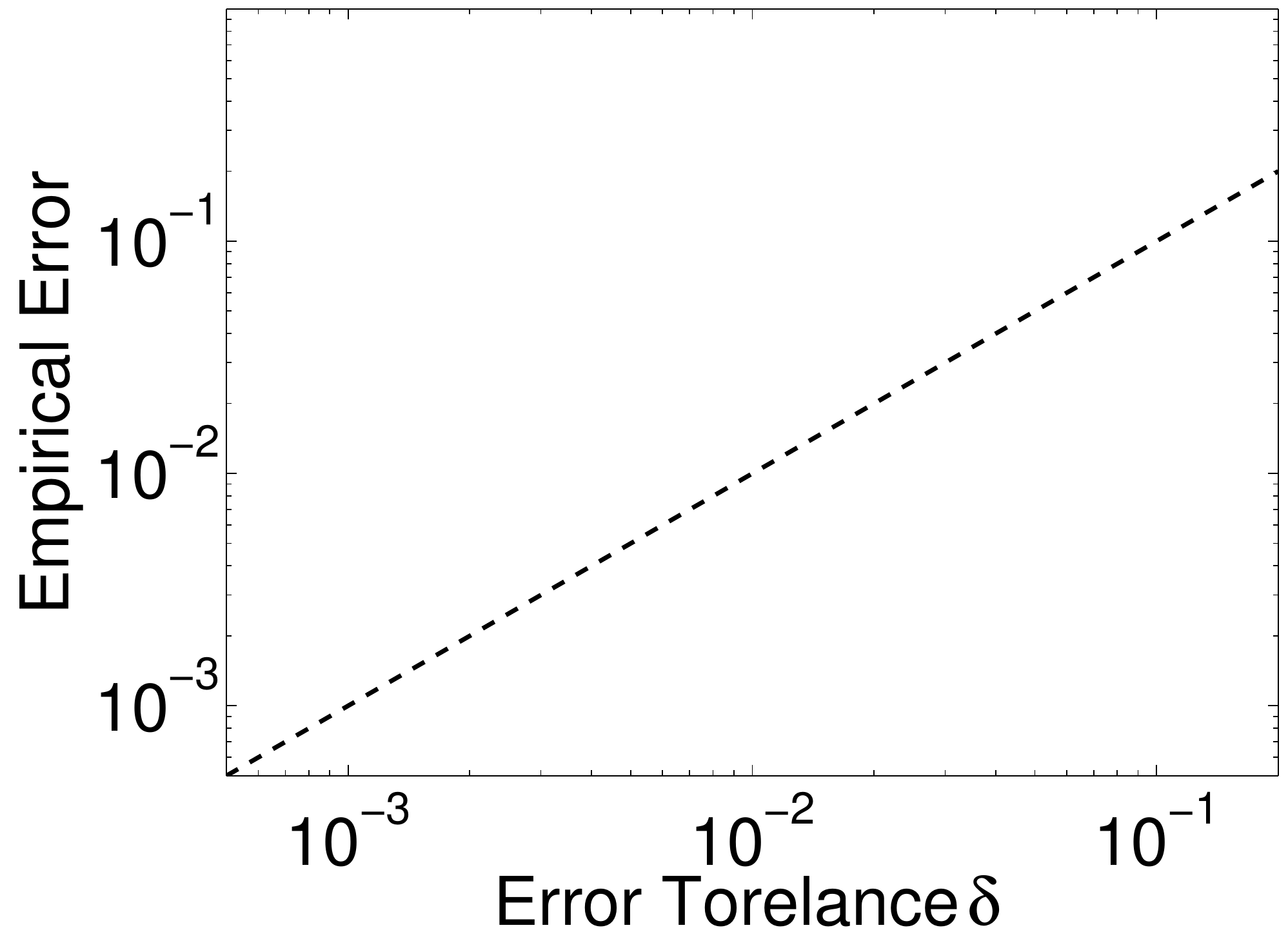}%
  }%
  ~
  \subfigure[$\sg=10^{-5}$, very easy]{%
    \label{fig:toy_ms_error_3}%
    \includegraphics[width=0.32\textwidth]{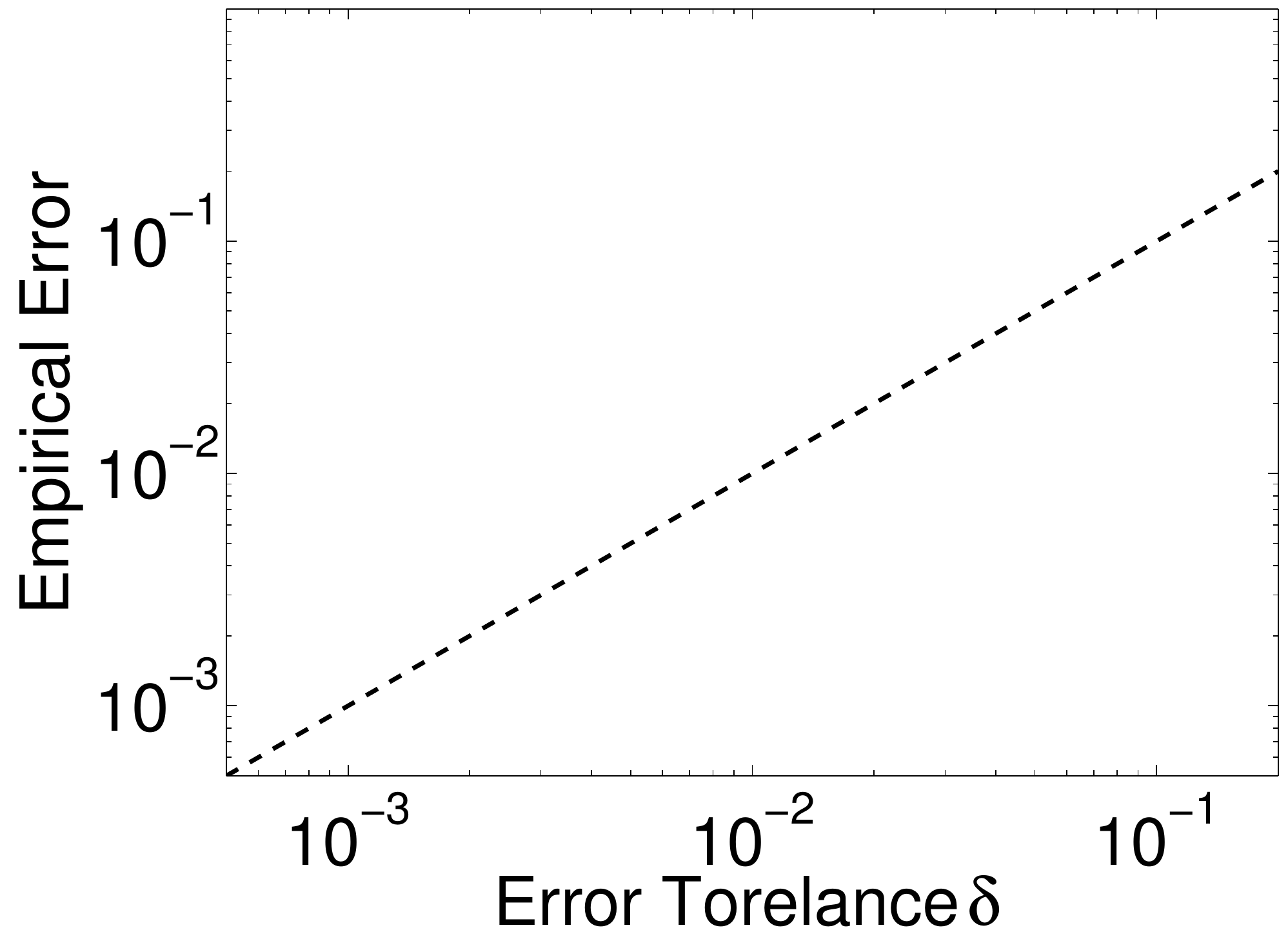}%
  }%
  \\
  \subfigure[$\sg=0.1$]{%
    \label{fig:toy_ms_data_1}%
    \includegraphics[width=0.32\textwidth]{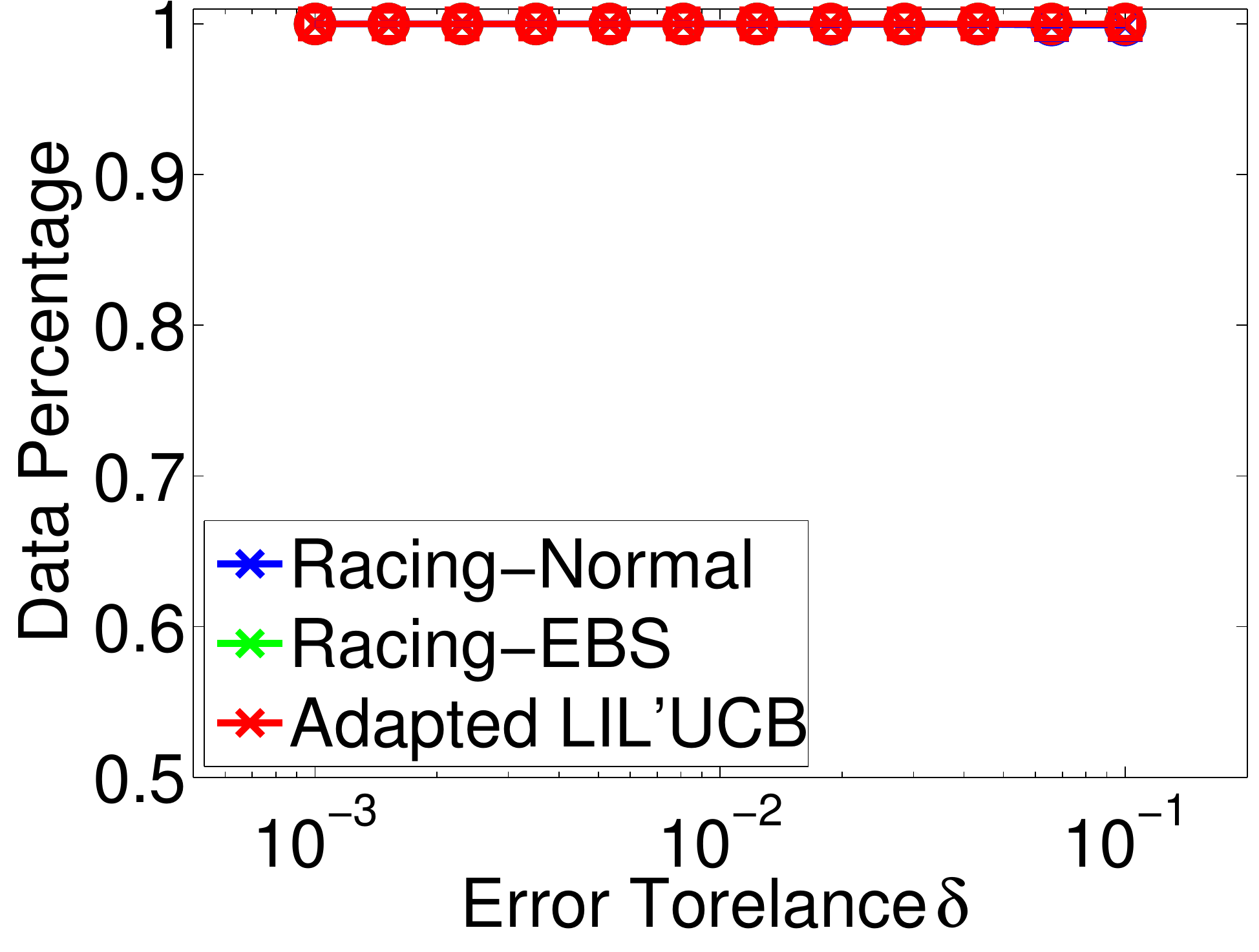}%
  }%
  ~
  \subfigure[$\sg=10^{-4}$, in log scale]{%
    \label{fig:toy_ms_data_2}%
    \includegraphics[width=0.32\textwidth]{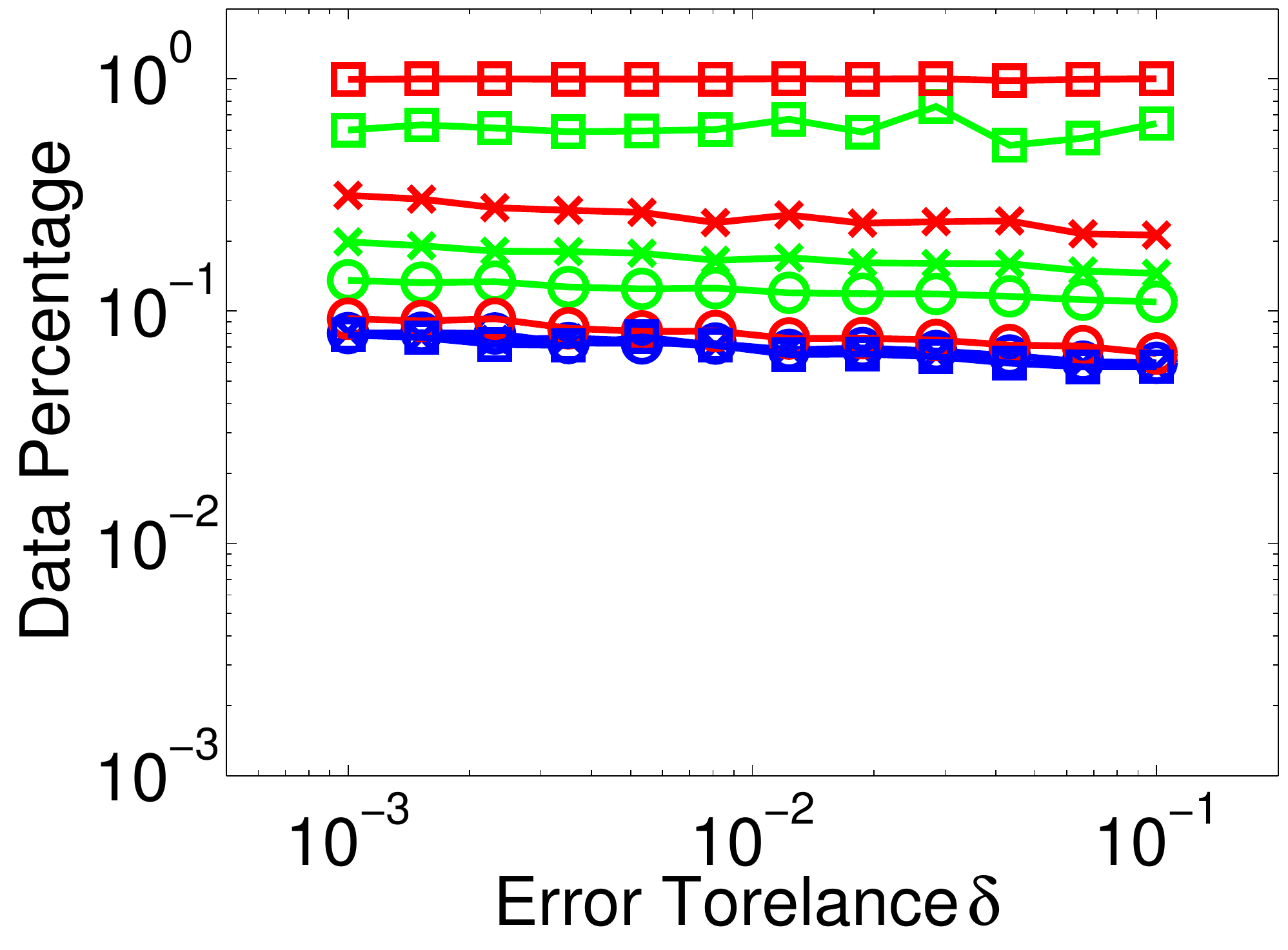}%
  }%
  ~
  \subfigure[$\sg=10^{-5}$, in log scale]{%
    \label{fig:toy_ms_data_3}%
    \includegraphics[width=0.32\textwidth]{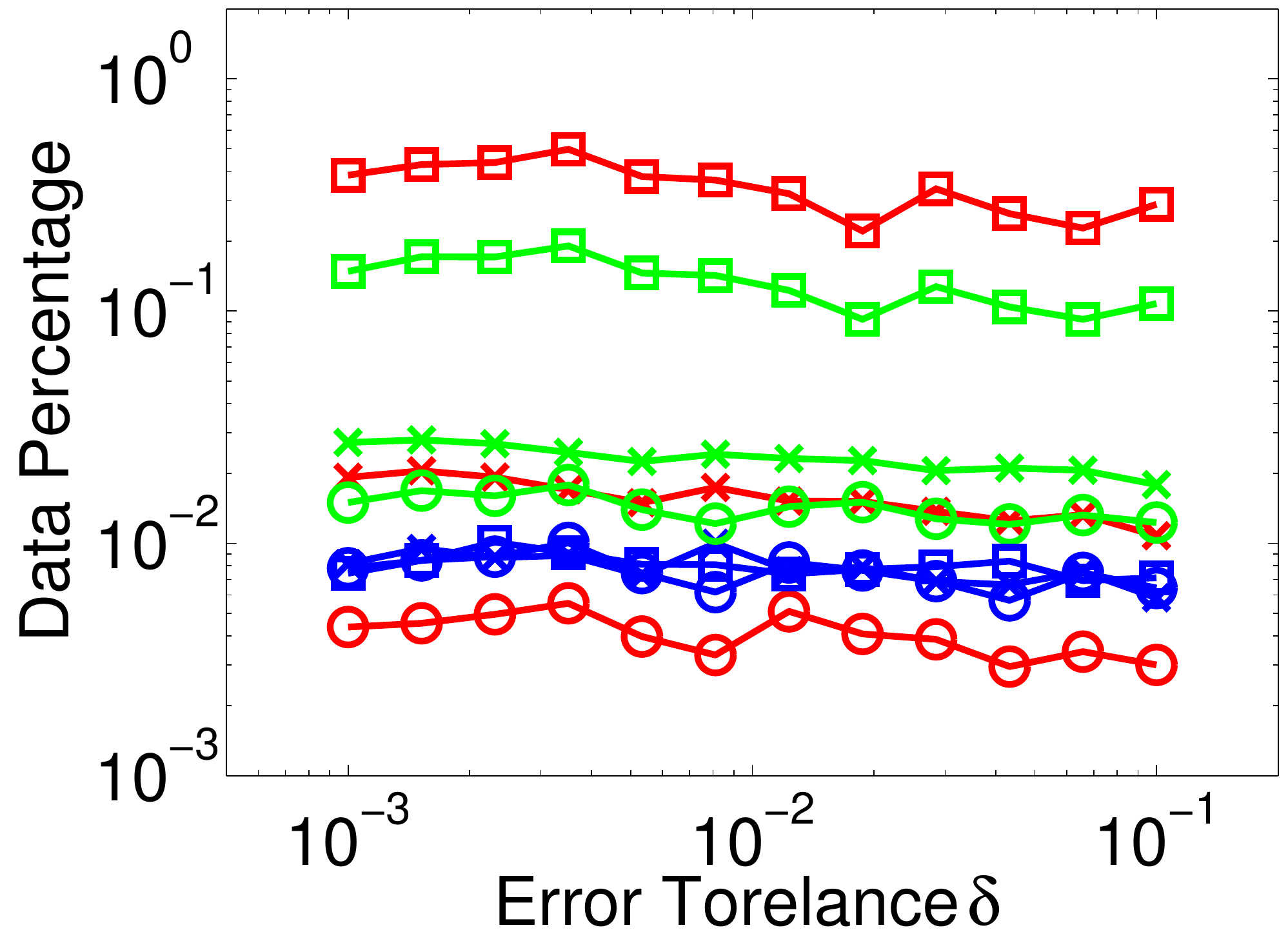}%
  }%
  \caption{Synthetic data. $D=10$. Racing uses marginal variance estimate $\hat{\sg}_i$. (\subref{fig:toy_ms_error_1},\subref{fig:toy_ms_error_2},\subref{fig:toy_ms_error_3}) Estimated error with $95\%$ confidence interval. Plots not shown if no error occured. (\subref{fig:toy_ms_data_1},\subref{fig:toy_ms_data_2},\subref{fig:toy_ms_data_3}) proportion of sampled data. $\log f_n(i)$ is sampled from Normal ($\times$), Uniform ($\bigcirc$) and LogNormal ($\square$) distributions. Plots of Racing-Normal overlap in (\subref{fig:toy_data_1},\subref{fig:toy_data_2},\subref{fig:toy_data_3}).}\label{fig:toy_ms}%
\end{figure*}

The results with the marginal variance estimate $\hat{\sg}_i$ for Racing are shown in Fig.~\ref{fig:toy_ms}. The Racing algorithms (both EBS and Normal) performs more conservatively compared to the plots when using pairwise variance estimate $\hat{\sg}_{i,j}$ in Fig.~\ref{fig:toy}, but the relative performance of all the algorithms are very similar to Fig.~\ref{fig:toy}. 

\begin{figure*}[p]
\centering
\begin{minipage}[t]{\textwidth}%
  \centering
  \subfigure[$\sg=0.1$, very hard]{%
    \label{fig:toy_D2_error_1}%
    \includegraphics[width=0.32\textwidth]{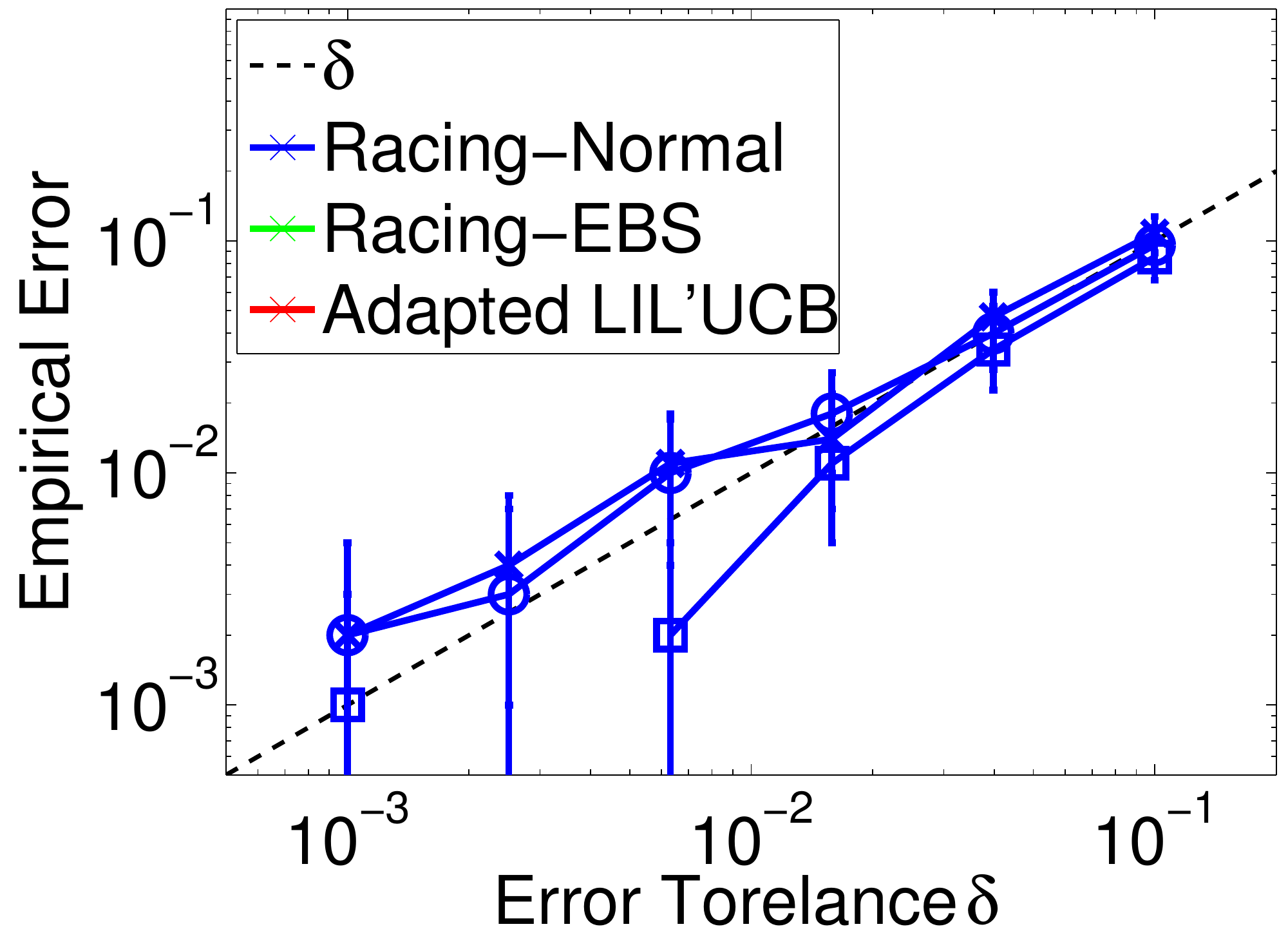}%
  }%
  ~
 \subfigure[$\sg=10^{-4}$, easy]{%
    \label{fig:toy_D2_error_2}%
    \includegraphics[width=0.32\textwidth]{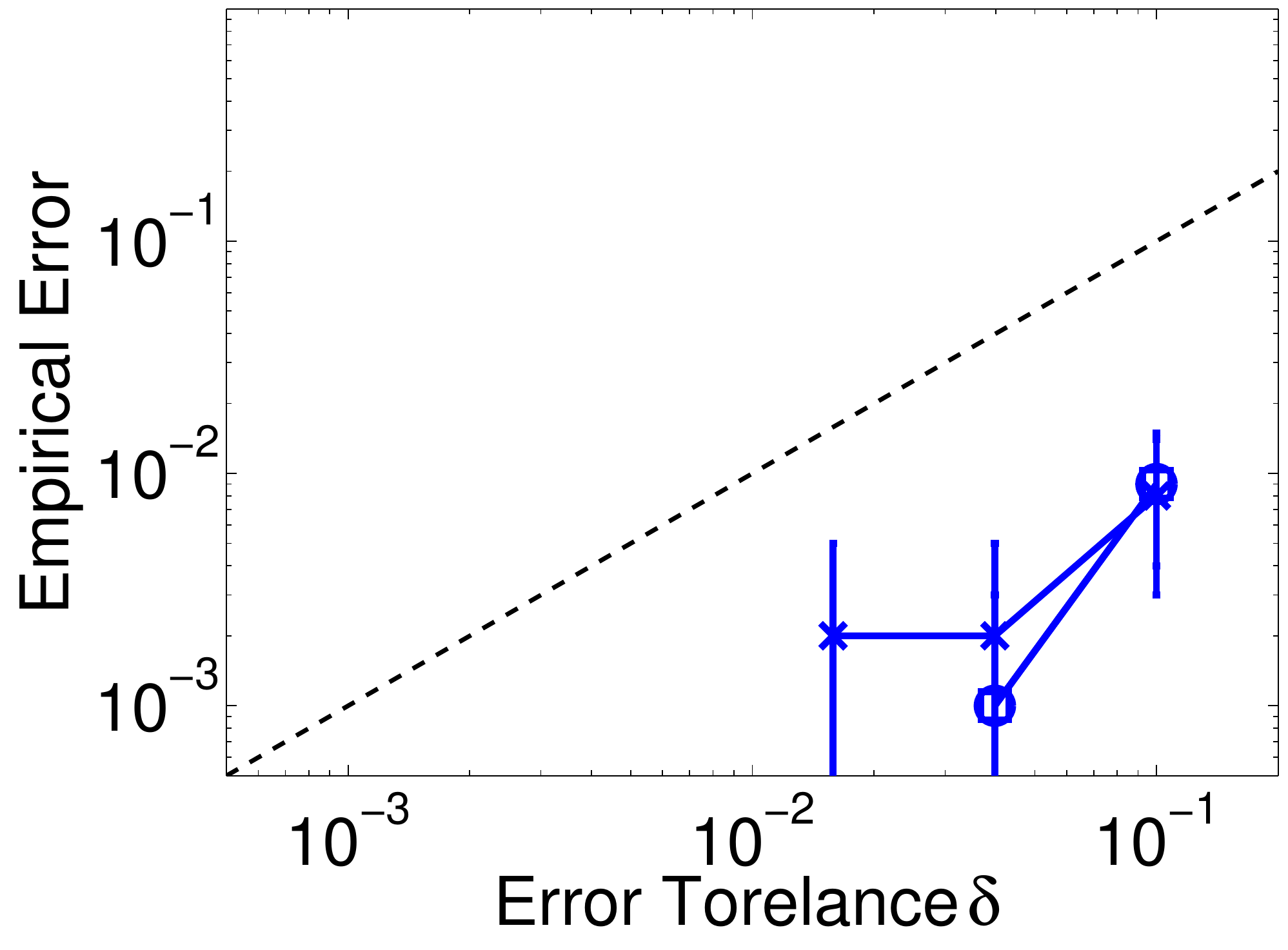}%
  }%
  ~
  \subfigure[$\sg=10^{-5}$, very easy]{%
    \label{fig:toy_D2_error_3}%
    \includegraphics[width=0.32\textwidth]{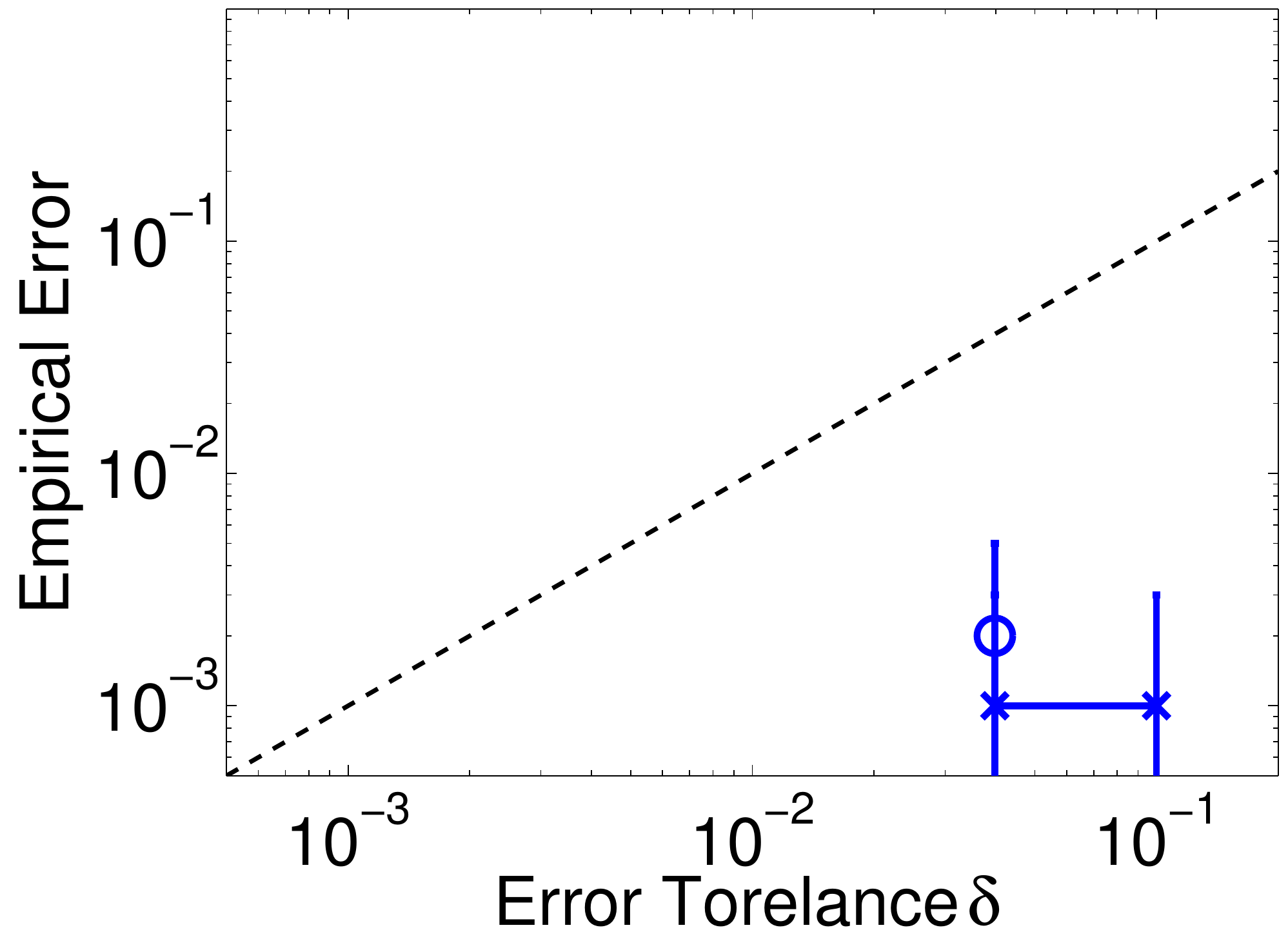}%
  }%
  \\
  \subfigure[$\sg=0.1$]{%
    \label{fig:toy_D2_data_1}%
    \includegraphics[width=0.32\textwidth]{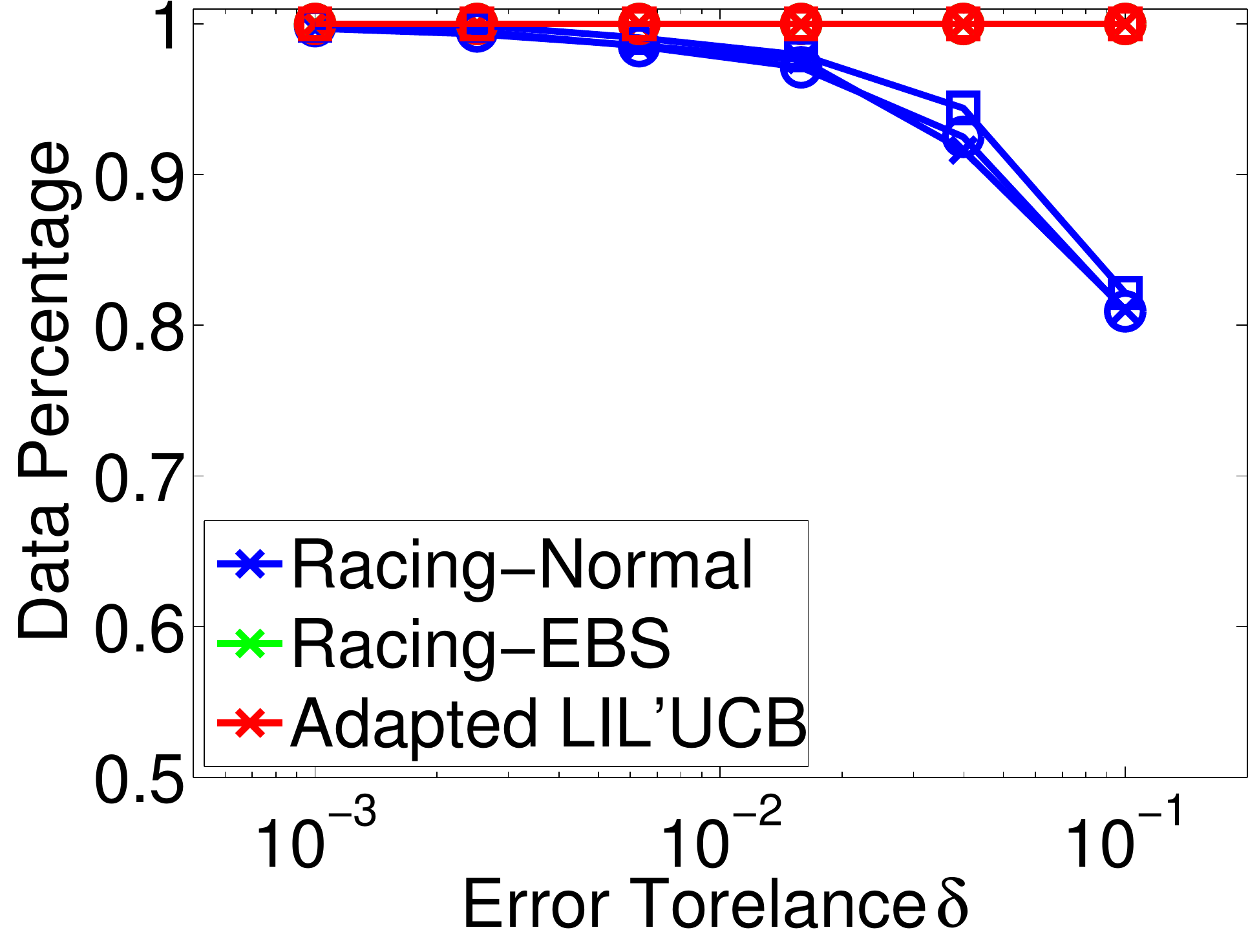}%
  }%
  ~
  \subfigure[$\sg=10^{-4}$, in log scale]{%
    \label{fig:toy_D2_data_2}%
    \includegraphics[width=0.32\textwidth]{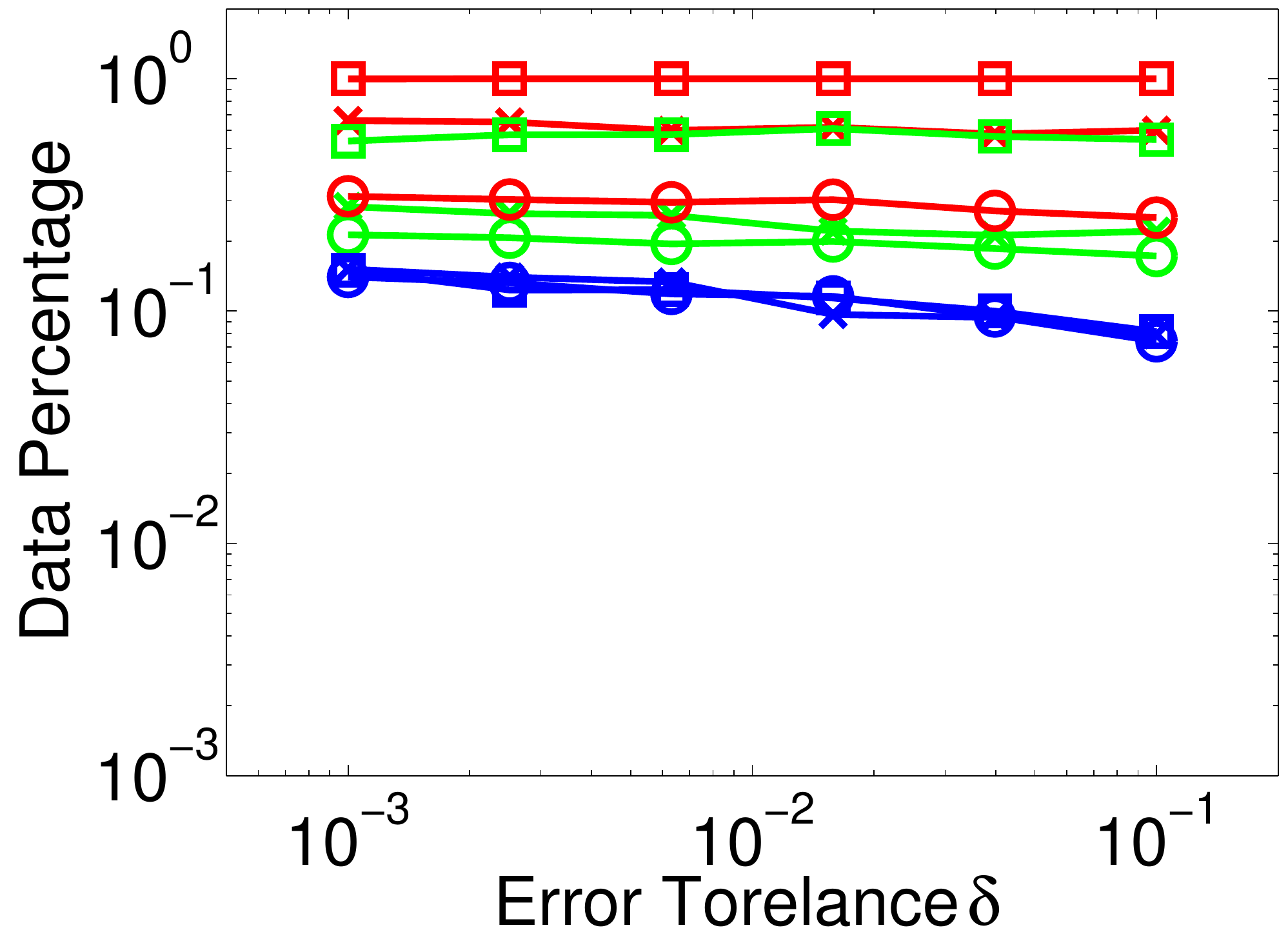}%
  }%
  ~
  \subfigure[$\sg=10^{-5}$, in log scale]{%
    \label{fig:toy_D2_data_3}%
    \includegraphics[width=0.32\textwidth]{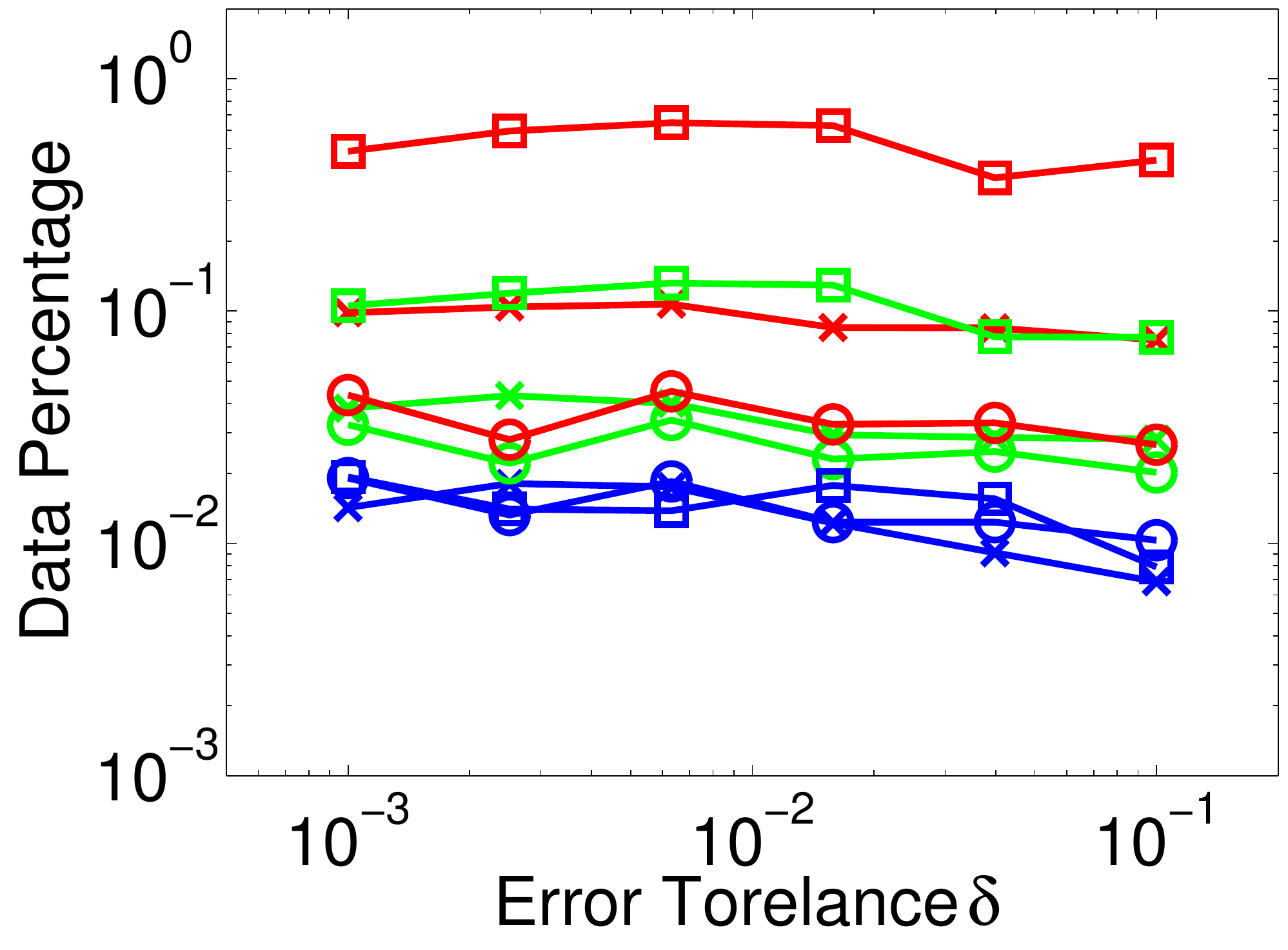}%
  }%
  \caption{Synthetic data. $D=2$. Racing uses pairwise variance estimate $\hat{\sg}_{i,j}$. (\subref{fig:toy_D2_error_1},\subref{fig:toy_D2_error_2},\subref{fig:toy_D2_error_3}) Estimated error with $95\%$ confidence interval. Plots not shown if no error occured. (\subref{fig:toy_D2_data_1},\subref{fig:toy_D2_data_2},\subref{fig:toy_D2_data_3}) proportion of sampled data. $\log f_n(i)$ is sampled from Normal ($\times$), Uniform ($\bigcirc$) and LogNormal ($\square$) distributions. Plots of Racing-Normal overlap in (\subref{fig:toy_data_1},\subref{fig:toy_data_2},\subref{fig:toy_data_3}).}\label{fig:toy_D2}%
\end{minipage}
\\
\begin{minipage}[t]{\textwidth}
  \centering
  \subfigure[$\sg=0.1$, very hard]{%
    \label{fig:toy_D100_error_1}%
    \includegraphics[width=0.32\textwidth]{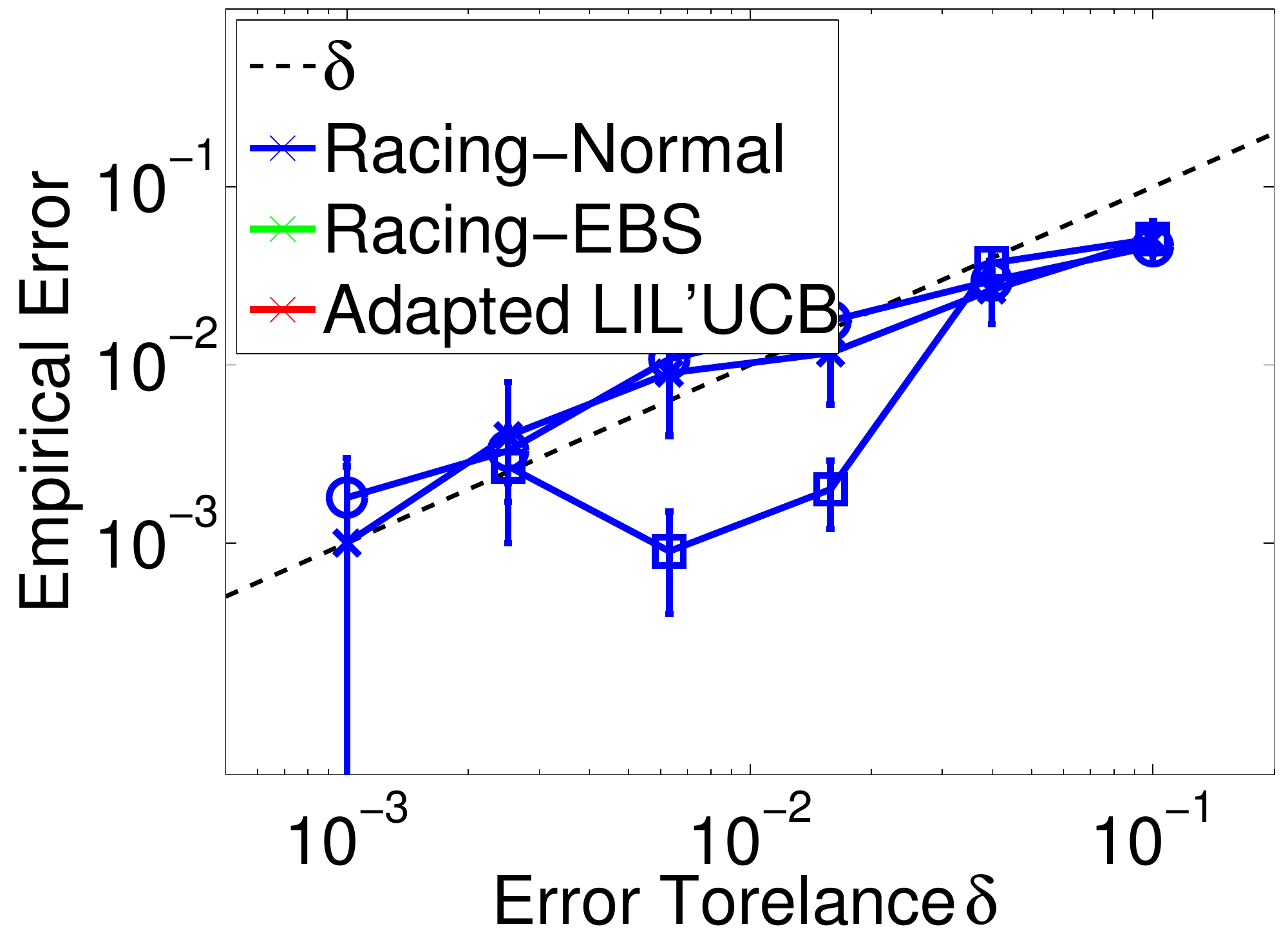}%
  }%
  ~
 \subfigure[$\sg=10^{-4}$, easy]{%
    \label{fig:toy_D100_error_2}%
    \includegraphics[width=0.32\textwidth]{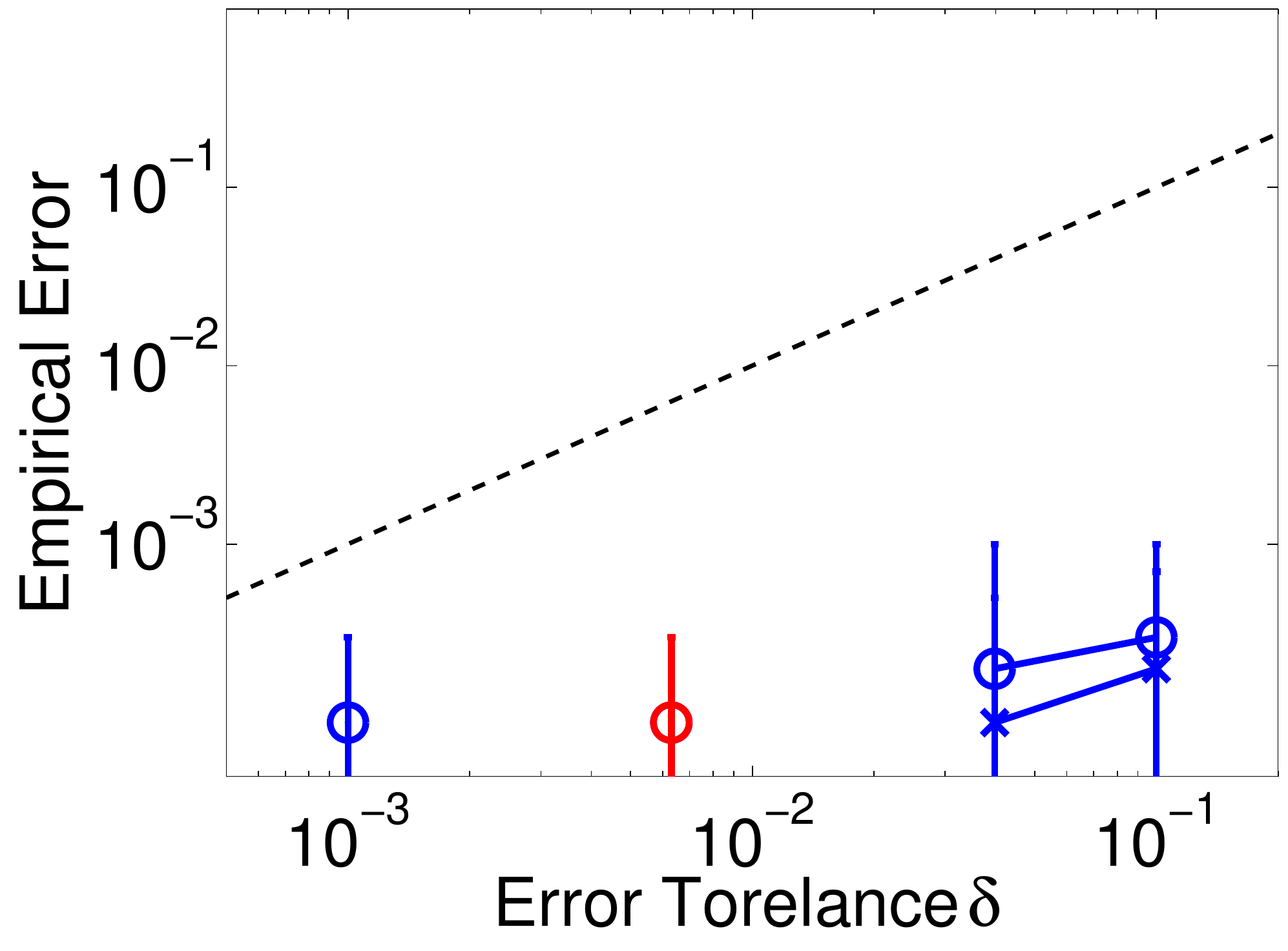}%
  }%
  ~
  \subfigure[$\sg=10^{-5}$, very easy]{%
    \label{fig:toy_D100_error_3}%
    \includegraphics[width=0.32\textwidth]{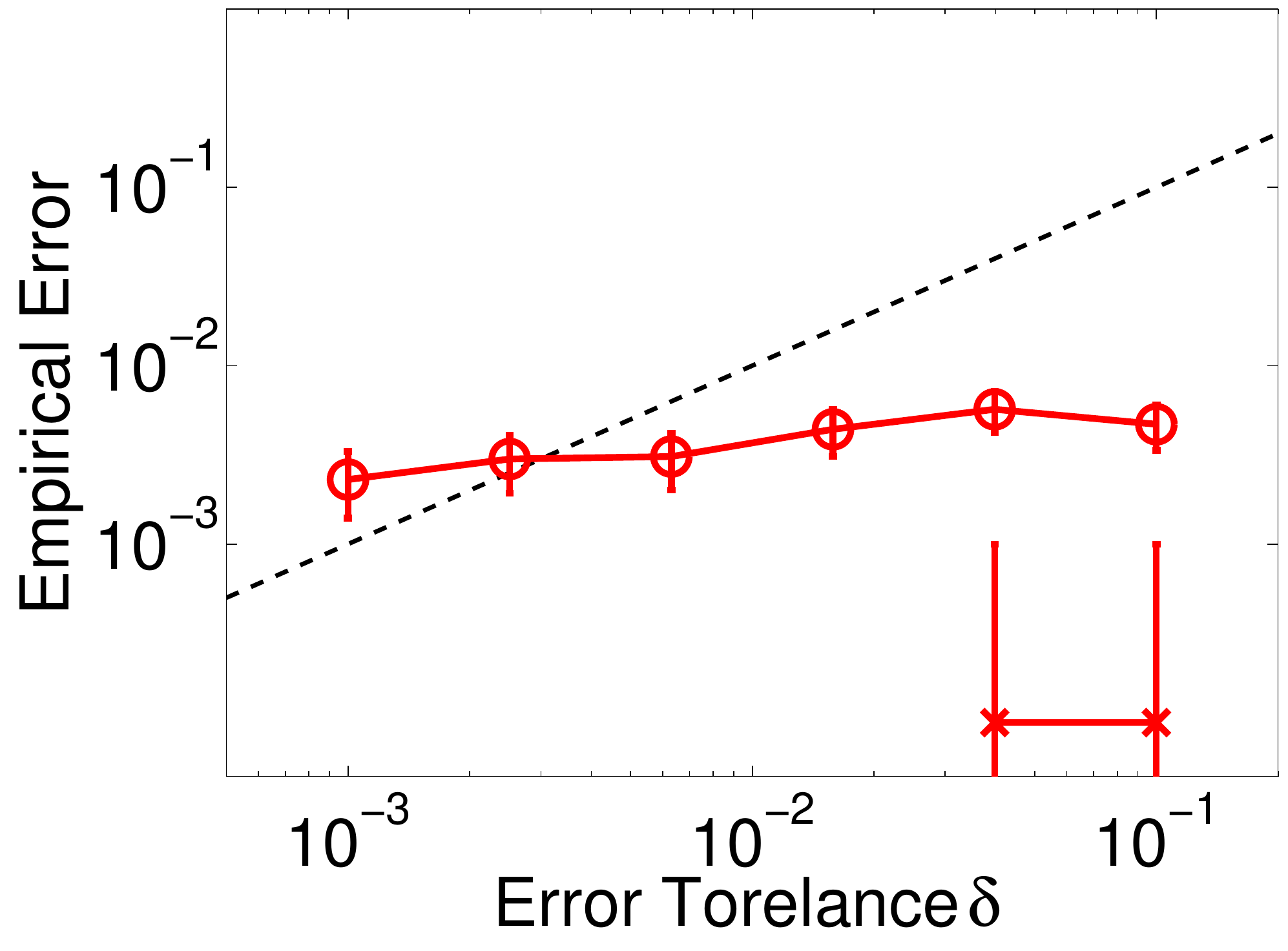}%
  }%
  \\
  \subfigure[$\sg=0.1$]{%
    \label{fig:toy_D100_data_1}%
    \includegraphics[width=0.32\textwidth]{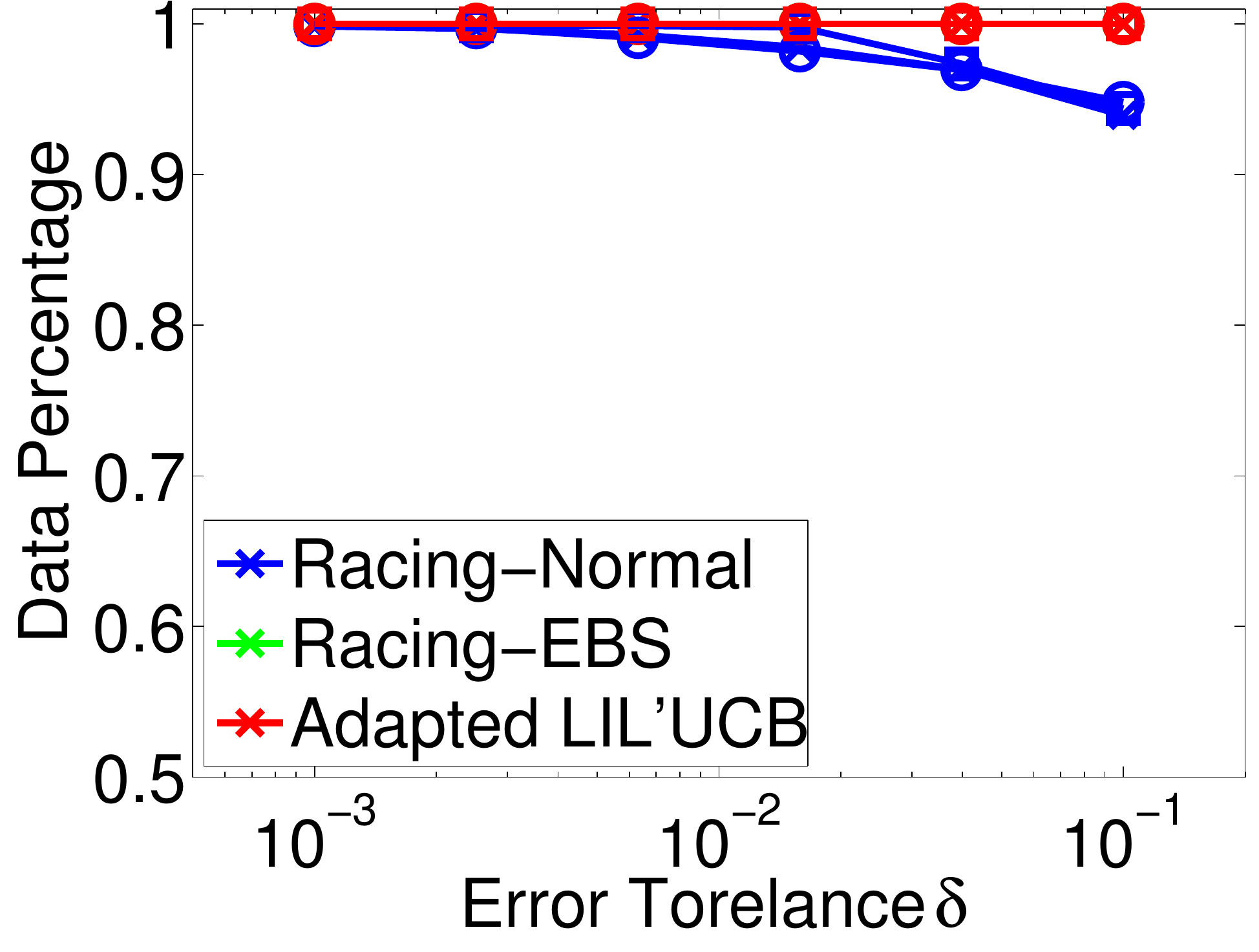}%
  }%
  ~
  \subfigure[$\sg=10^{-4}$, in log scale]{%
    \label{fig:toy_D100_data_2}%
    \includegraphics[width=0.32\textwidth]{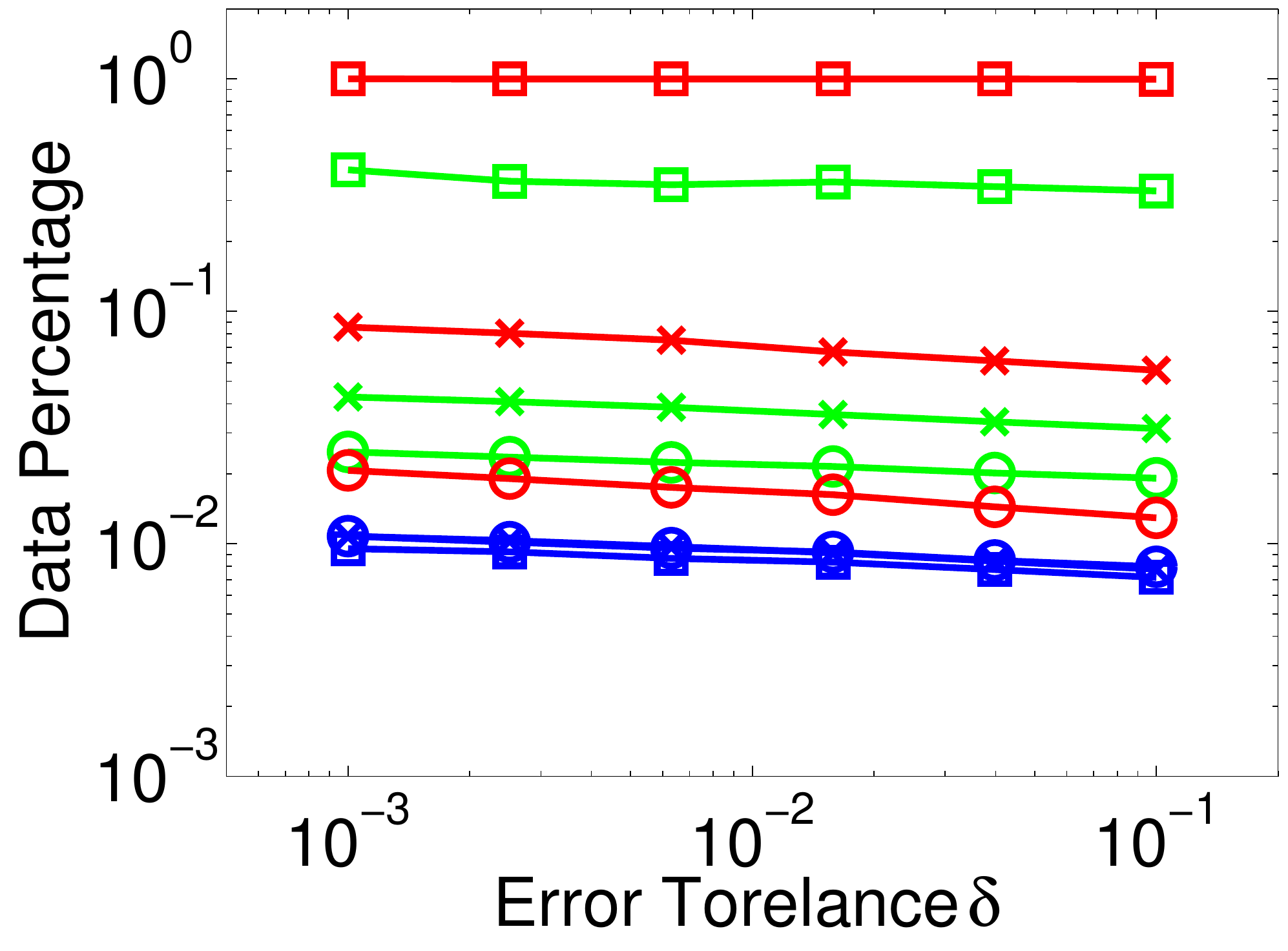}%
  }%
  ~
  \subfigure[$\sg=10^{-5}$, in log scale]{%
    \label{fig:toy_D100_data_3}%
    \includegraphics[width=0.32\textwidth]{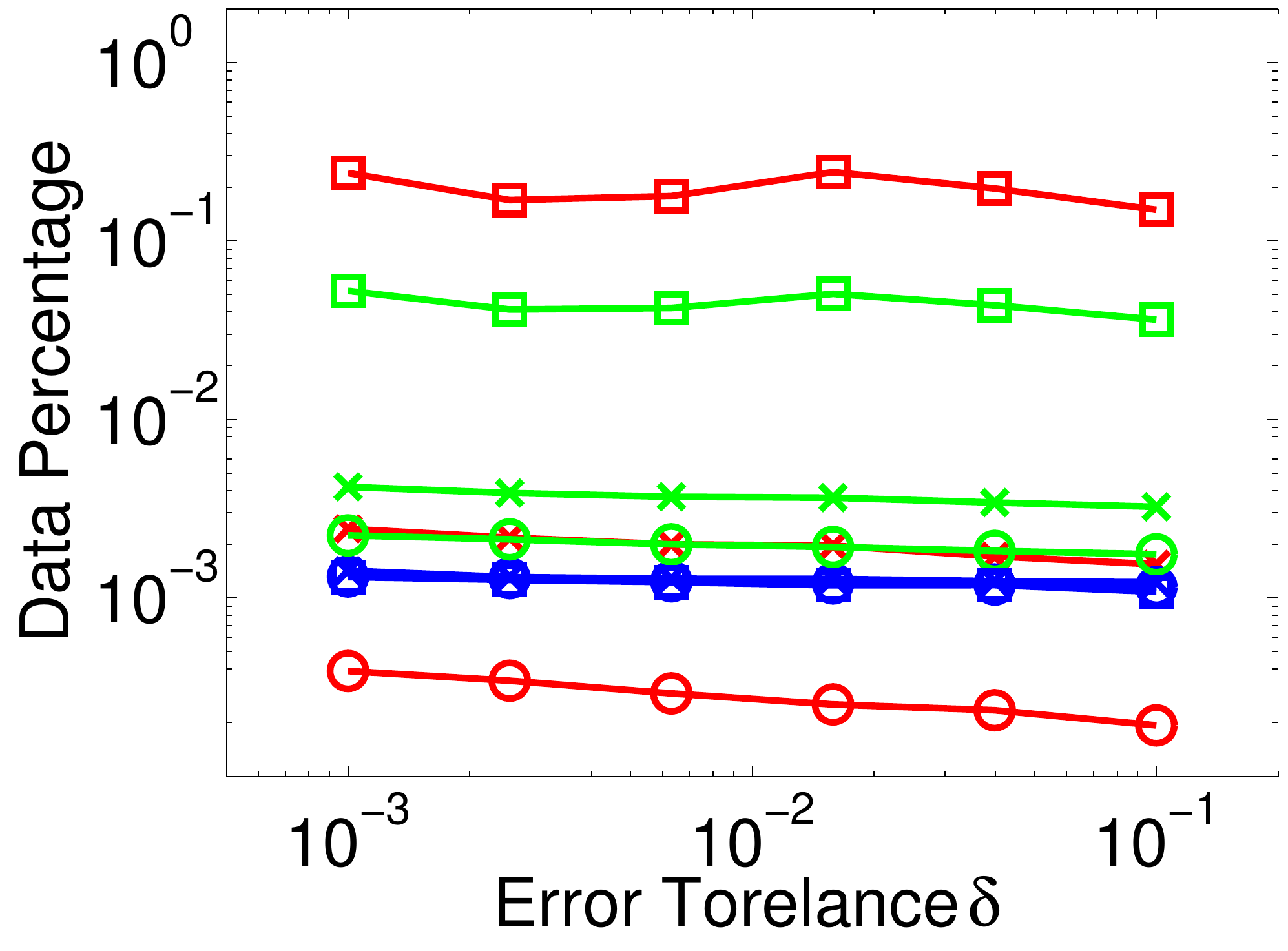}%
  }%
  \caption{Synthetic data. $D=100$. Racing uses pairwise variance estimate $\hat{\sg}_{i,j}$. (\subref{fig:toy_D100_error_1},\subref{fig:toy_D100_error_2},\subref{fig:toy_D100_error_3}) Estimated error with $95\%$ confidence interval. Plots not shown if no error occured. (\subref{fig:toy_D100_data_1},\subref{fig:toy_D100_data_2},\subref{fig:toy_D100_data_3}) proportion of sampled data. $\log f_n(i)$ is sampled from Normal ($\times$), Uniform ($\bigcirc$) and LogNormal ($\square$) distributions. Plots of Racing-Normal overlap in (\subref{fig:toy_data_1},\subref{fig:toy_data_2},\subref{fig:toy_data_3}).}\label{fig:toy_D100}%
\end{minipage}%
\end{figure*}

We also provide the results with $D=2$ and $D=100$ when Racing algorithms use pairwise variance estimate in Fig.~\ref{fig:toy_D2} and \ref{fig:toy_D100} respectively. Racing-Normal performs the best in all situations and the empirical error never exceeds the provided bound $\de$ with a statistical significance of $0.05$. 

Notice that the error of adaptive lil'UCB exceeds the error tolerance in the experiment with $D=100$ and $l_{i,n}\sim \mathrm{Uniform}[0,1]$. This is because we use the recommended heuristic setting of parameters in \citet{jamieson2014lil} that unfortunately does not satisfy the theoretical guarantee of Thm.\ 2 in \citet{jamieson2014lil}. lil'UCB (heuristic) performed significantly better than the setting with guarantees in \citet{jamieson2014lil}. So we expect that adaptive lil'UCB with parameters satisfying Thm.\ 2 of \citet{jamieson2014lil} will perform significantly worse than adaptive lil'UCB (heuristic) and Racing-Normal in terms of the reward sample complexity.

\subsection{Details of the Bayesian ARCH Model Selection Experiment}\label{sec:extra_exp_arch}
An ARCH model is commonly used to model the stochastic volatility of financial times series. Let $r_t\defeq \log (p_t/p_{t-1})$ be the logarithm return of some asset price $p_t$ at time $t$. We assume a constant mean process for the return and remove the estimated mean in a pre-process step. An important problem in applying ARCH for financial data is to choose the complexity, the order $q$ of the auto-regressive model. We treat the model selection problem as a Bayesian inference problem for the random variable $q$. We use a uniform prior distribution, $\pi(q) = 1 / |\mathbb{Q}|$.

An MCMC algorithm was introduced in \citet{CarlinChib95} to infer the posterior model distribution by augmenting the parameter space to a complete parameter set for all models $((\alpha_i^{(j)})_{i=0}^j, \nu^{(j)}), j\in \mathbb{Q}$, then assigning the regular prior for the selected model $j=q$ and pseudopriors for those models that are not selected $j\neq q$. Then regular MCMC algorithms can be applied to sample all the random variables $q, ((\al_i^{(j)})_i, \nu^{(j)})_j$ without the problem of transdimensional moves as in reversible jump MCMC.

The mixing rate of \citet{CarlinChib95} depends on a proper choice of the pseudoprior for $(\alpha_i^{(j)}, \nu^{(j)})$. Ideally it should be similar to the parameter posterior when the model is chosen $p(\alpha_i^{(j)}, \nu^{(j)})|q=j, \br)$. We first reparameterize $(\al_i^{(j)}, \nu^{(j)})$ with a softplus function $x=\log(1+\exp(x'))$ to allow a full support along the real axis and then take the Laplace approximation at the MAP of transformed parameters as the pseudoprior for each model separately.

In order to avoid accessing the entire dataset each iteration, we use subsampling-based algorithms to sample all the conditionals except the pseudoprior as follows
\begin{align}
q | (\bal^{(j)}, \nu^{(j)})_j &\sim \pi(q) \prod_t p(r_t|\bal^{(q)}, \br_{t-q:t-1}, \nu^{(q)}),\nn\\
(\bal^{(q)}, \nu^{(q)}) | q &\sim p(\bal^{(q)})p(\nu^{(q)}) \prod_t p(r_t|\bal^{(q)}, \br_{t-q:t-1}, \nu^{(q)}),\nn\\
(\bal^{(j)}, \nu^{(j)}) | q &\overset{iid}\sim p_{\mathrm{pseudoprior}}(\bal^{(j)},\nu^{(j)}), \forall j \neq q,
\end{align}
where we sample $q$ with Racing-Normal Gibbs and sample $\bal^{(q)}, \nu^{(q)}$ using MH with a proposal from SGLD and a rejection step provided by Racing-Normal MH. The rejection step controls the error introduced in SGLD when the step size is large.

As the marginal likelihood for each model could be differed by a few orders of magnitudes, to make sure every model is sampled sufficiently often, we first adjust the prior distribution $\tilde{\pi}$ with the Wang-Landau algorithm with an annealing adaptation on $\log \tilde{\pi}$, $1/(1+t/100)$, so that the posterior distribution $\tilde{p}(q|\br)$ is approximately uniform. We then fix $\tilde{\pi}$ and compare the exact and approximate MCMC algorithms. The real posterior distribution can be computed as $p(q|\br) \propto \tilde{p}(q|\br)/\tilde{\pi}(q)$.

We choose the step size separately for the exact and stochastic gradient Langevin dynamics \cite{welling2011bayesian} so that the acceptance rate is about 36\%.

We apply the control variates by first segmenting the 2-D space of $\bz_{j,t}\defeq (r_t, \al_0^{(j)}+(\bal_{1:j}^{(j)})^T \br_{t-j:t-1})$, where $\bal^{(j)}$ takes the MAP value, equally into 100 bins according to marginal quantiles and then taking the reference points at the mean of each bin. We also notice that some data points have large residual reward $l_{i,n}-h_{i,n}$ when $\bz_{j,t}$ is far from the reference point. We take 20\% of the points with the largest distance in $\bz$ as outliers, always compute them every iteration and apply the subsampling algorithm for the rest data.

\subsection{Details of the Author Coreference Experiment}\label{sec:extra_exp_rexa}
The main differences of this sampling problem from Eq.~\ref{eq:tilde_p} are that 
\begin{enumerate}
\item $|C_y|\neq |C_{y'}|$ and the distribution of the cluster size follows approximately a power law with the value varying from as small as 1 to thousands. If we set $m^{(1)}=50$ as usual, we already draw about 33\% of all the rewards in the first mini-batch. So we slightly abuse the Normal assumption and use a small size for $m^{(1)}=3$ and use doubling scheme for the rest with $m_y^{(2)}= (|C_y| - 3) / 10 \wedge 1$. The experiment shows an empirical error $0.045$ of mis-identification of the best arm with the provided bound $\de=0.05$.
\item The distribution of $\{f_{\ta}(x_i,x_j):j\in C_y\}$ is independent from different clusters/arms. We exploit the independence of rewards and choose the bound
\begin{align}
&G_{\mathrm{Normal}}(\de,T_i,T_j,\hat{\sg}_i,\hat{\sg}_j)\nn\\
&=\left(\frac{\hat{\sg}_i}{T_i}\left(1-\frac{T_i-1}{N_i-1}\right)+\frac{\hat{\sg}_j}{T_j}\left(1-\frac{T_j-1}{N_j-1}\right)\right)^{-1/2}B_{\mathrm{Normal}}.
\end{align}
This modification has the same performance as with the pairwise variance estimate and has the same computational complexity as with the marginal variance estimate $\mathcal{O}(DN)$.
We compute $B_{\mathrm{Normal}}$ with a sub-optimal but simpler choice as
\begin{equation}
B_{\mathrm{Normal}}(\de) = \Phi^{-1}\left(1 - \frac{\de}{t^*-1}\right). \label{eq:B_normal_2}
\end{equation}
It is easy to show that Eq.~\ref{eq:condition_G} still holds in this case using a union bound across $t$. The bound in Eq.~\ref{eq:B_normal_2} is strictly looser than $B_{\mathrm{Normal}}=\cE^{-1}(\de)$ but the difference is small when $\de\ll 1$ and diminishes to $0$ as $\de\rightarrow 0$.
\end{enumerate}
We obtained the dataset from the authors of \citet{singh2012monte} but it is different from what is used in \citet{singh2012monte} with more difficult citations. The best $B^3$ F-1 score reported in this paper is a reasonable value for this data set according to personal communications with the authors of \citet{singh2012monte}.

\end{document}